\pdfoutput=1

\documentclass[11pt]{article}

\usepackage[final]{acl}

\usepackage{times}
\usepackage{latexsym}

\usepackage{algpseudocode} 
\usepackage{algorithm} 
\usepackage{multirow}

\usepackage[T1]{fontenc}

\usepackage[utf8]{inputenc}

\usepackage{microtype}

\usepackage{inconsolata}

\usepackage{graphicx}
\usepackage{subcaption}

\newcommand{\mistral}{Mistral-7B}
\newcommand{\qwensmall}{Qwen-7B}
\newcommand{\qwenbig}{Qwen-72B}
\newcommand{\aya}{Aya-Expanse-32B}
\newcommand{\llamasmall}{Llama 3-8B}
\newcommand{\llamabig}{Llama 3-70B}
\newcommand{\gptmini}{GPT-4o-mini}

\definecolor{darkgreen}{rgb}{0.0, 0.5, 0.0}
\definecolor{burgundy}{rgb}{0.5, 0.0, 0.13}
\title{Analyzing Political Bias in LLMs via Target-Oriented Sentiment Classification}

\author{
 \textbf{Akram Elbouanani\textsuperscript{1}},
 \textbf{Evan Dufraisse\textsuperscript{1}},
 \textbf{Adrian Popescu\textsuperscript{1}},
\\
\\
 \textsuperscript{1}Université Paris-Saclay, CEA, List, F-91120, Palaiseau, France
}

\begin{document}

\maketitle
\begin{abstract}






Political biases encoded by LLMs might have detrimental effects on downstream applications. 
Existing bias analysis methods rely on small-size intermediate tasks (questionnaire answering or political content generation) and rely on the LLMs themselves for analysis, thus propagating bias. 
We propose a new approach leveraging the observation that LLM sentiment predictions vary with the target entity in the same sentence. 
We define an entropy-based inconsistency metric to encode this prediction variability. 
We insert 1319 demographically and politically diverse politician names in 450 political sentences and predict target-oriented sentiment using seven models in six widely spoken languages. 
We observe inconsistencies in all tested combinations and aggregate them in a statistically robust analysis at different granularity levels.
We observe positive and negative bias toward left and far-right politicians and positive correlations between politicians with similar alignment.
Bias intensity is higher for Western languages than for others.
Larger models exhibit stronger and more consistent biases and reduce discrepancies between similar languages. 
We partially mitigate LLM unreliability in target-oriented sentiment classification (TSC) by replacing politician names with fictional but plausible counterparts.
\end{abstract}

\begin{figure}[ht]
    \centering
    \includegraphics[width=0.99\linewidth]{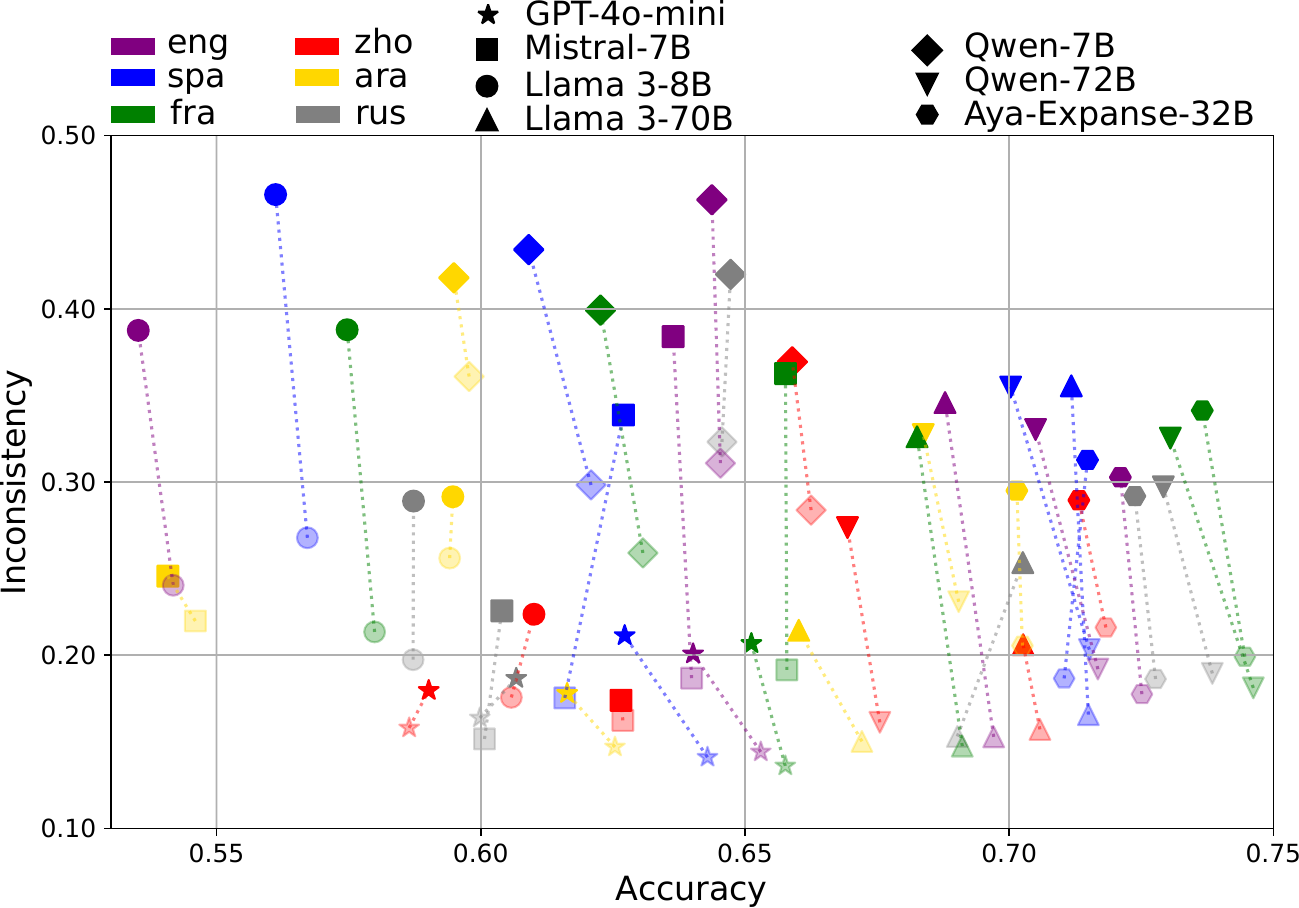}
    \caption{Sentiment-prediction based analysis of LLM model--language combinations when varying names in political sentences before and after politician name replacement with fictional but plausible names (no vs. 0.7 transparency). \textbf{The desired behavior combines high accuracy, reflecting a correct understanding of the sentiment associated with names, and low inconsistency (Eq.~\ref{eq_consistency}), reflecting a lack of bias toward the analyzed entities}. The comparison highlights the entity-related bias encoded in LLMs and the effectiveness of the name replacement mitigation approach. }
    \label{fig_teaser}
\end{figure}

\section{Introduction}
Large Language Models (LLMs) have reshaped the AI landscape due to their versatility and ease of use.
Beyond their generation capabilities, they greatly simplified NLP tasks that previously required strong expertise and specific development~\cite{kojima2022large,qin2024large}.
However, multiple biases, including demographic~\cite{salinas2023unequal} and political ones~\cite{motokiMoreHumanHuman2024}, question their use in socially important real-life applications, including the moderation of political discussions on social media platforms~\cite{kolla2024llm} or the automatic analysis of media coverage of political events or discourse~\cite{hsu-etal-2024-enhancing}.
Addressing these tasks requires components such as hate speech detection, stance detection, or target-oriented sentiment analysis. 
These components entail subjectiveness and are bias-prone, making the understanding and mitigation of this phenomenon crucial for ensuring accuracy, fairness, and reliability.  

Existing LLM political bias studies fall into two main categories.
Questionnaire-based works~\cite{hartmannPoliticalIdeologyConversational2023,motokiMoreHumanHuman2024} repurpose methodologies, such as the Political Compass~\cite{eysenck2018psychology}, to infer LLM political leanings based on answers to predefined questions. 
Generation-based methods~\cite{bangMeasuringPoliticalBias2024,buylLargeLanguageModels2024,huangReducingSentimentBias2020} prompt LLMs to generate political content and deploy sentiment or stance detection to infer ideological alignments. 
While interesting, these approaches have significant limitations.
First, the number of interactions with LLMs is usually reduced, leading to limited statistical power.
Second, they depend on the prompt wording, with high result variability occurring even after minimal changes. 
Third, they provide limited actionable insights about LLM use in downstream applications. 

Informed by these limitations, we propose a new political bias analysis framework leveraging LLM prediction inconsistency in target-oriented sentiment classification.
Assume we apply TSC to:  
``\textit{$\mathcal{X}$ has limited means but contributed to solving this conflict.}''
A reliable classifier should consistently predict the same sentiment for any name replacing $\mathcal{X}$. 
We define an entropy-based inconsistency metric to quantify this undesired prediction variability. 
To verify this hypothesis, we use 450 political sentences from MAD-TSC~\cite{dufraisse-etal-2023-mad}, an existing dataset, and 1319 politicians representing diversified demographics and eight political alignments.
We vary the names in the sentences, vary the models, and use sentence translations in six languages to obtain nearly 25 million data points and robustly analyze LLMs' political bias.

Figure~\ref{fig_teaser} analyzes political bias in LLMs using TSC accuracy and inconsistency metrics for 42 model--language combinations. 
Without mitigation, name-related inconsistency is significant for all models due to prediction variability when changing the politician names.
Confirming previous results~\cite{zhangSentimentAnalysisEra2024}, TSC accuracy is better for larger models. 
Section~\ref{sec-results} provides detailed results for political alignments, languages, models, and politicians.
We notably: 
(1) confirm positive and negative biases toward the left and the right political alignments, 
(2) find a positive correlation for politicians with a similar alignment,
(3) show that bias intensity varies with the language,
and 
(4) examine the effect of model size on biases. 
Replacing politicians with fictional but plausible names reduces inconsistency and slightly improves accuracy in most cases.
This replacement provides a simple yet effective way to reduce TSC biases. 
However, a degree of inconsistency subsists, questioning the reliability of LLM usage for NLP tasks involving subjectivity. 

\section{Related Work}


Sentiment analysis is a prominent NLP application \cite{ouyangShiftedOverlookedTaskoriented2023} used for analyzing media biases \cite{falckMeasuringProximityNewspapers2020, dufraisse2024combiningobjectivesubjectiveperspectives}, predicting stock market \cite{lopez-liraCanChatGPTForecast2023, wu2023bloomberggptlargelanguagemodel}, and detecting hate speech \cite{chiuDetectingHateSpeech2022}. 

Importantly, LLMs exhibit societal, cultural, and political biases, which manifest in generated text \cite{shengSocietalBiasesLanguage2021, shengWomanWorkedBabysitter2019, naousHavingBeerPrayer2024, 10.1093/pnasnexus/pgaf089}. 
These biases complicate sentiment analysis, particularly in politically sensitive contexts, as they can skew model predictions and reinforce preexisting biases \cite{unglessPotentialPitfallsAutomatic2023}.
We use these observations to highlight the predictions' unreliability and use them for political bias analysis. 

Studies investigating biases in LLMs use two main approaches. 
The first, \textbf{questionnaires and controlled tasks}, employs structured prompts or multiple-choice questions to measure biases. For instance, \cite{alkhamissiInvestigatingCulturalAlignment2024} use simulated sociological surveys to quantify cultural alignment, while \cite{santurkarWhoseOpinionsLanguage2023} leverage public opinion polls to evaluate alignment with demographic groups. \cite{hartmannPoliticalIdeologyConversational2023} and \cite{motokiMoreHumanHuman2024} utilize political compass tests and voting advice applications to assess political leanings. These methods provide easily quantifiable and reproducible results.
A key advantage of questionnaire-based approaches is their ability to systematically evaluate biases across diverse contexts, particularly when varying prompting strategies \cite{jiangCommunityLMProbingPartisan2022,abdulhaiMoralFoundationsLarge2024}.
However, they have significant limitations: constrained designs may not generalize to real-world applications \cite{lyuProbabilitiesUnveilingMisalignment2024}, and reliance on forced multiple-choice formats can misrepresent model behavior as well as not reflect real-world use cases \cite{wangMyAnswerFirstToken2024,rottgerPoliticalCompassSpinning2024}. 
Importantly, the questionnaire sizes are usually reduced~\cite{rozadoPoliticalPreferencesLLMs2024a}, making a robust statistical analysis of results difficult.

The second approach, \textbf{open-ended generation tasks}, involves prompting LLMs to generate essays, poems, or lengthy texts, for a less constrained analysis of biases. For example, \cite{gover2023political} apply sentiment analysis to political essays generated by GPT-3, revealing a moderate left-leaning bias. 
Similarly, \cite{mcgeeChatGptBiased2023} examine limericks generated by ChatGPT, uncovering patterns of bias favoring liberal politicians. These methods provide insights into how biases manifest in free-form text, but they face significant challenges. 
Open-ended generation offers a valuable complement to questionnaire-based approaches, particularly for capturing biases in context-rich scenarios \cite{buylLargeLanguageModels2024}.
However, quantifying biases in open-ended outputs is inherently difficult.
These studies often rely on word distribution analysis, sentiment or stance detection, or evaluation by LLMs as judges \cite{huangReducingSentimentBias2020,bangMeasuringPoliticalBias2024}, which are bias-prone themselves~\cite{unglessPotentialPitfallsAutomatic2023}. 
Moreover, the lack of control over output formats complicates bias assessment, as traditional algorithmic fairness metrics (e.g., equalized odds, demographic parity) are not directly applicable \cite{scherrer2024evaluating}. 
 
Sentiment analysis offers a complementary practical middle ground between structured questionnaires and open-ended text generation. Unlike rigid multiple-choice formats, it aligns with real-world language use while providing measurable bias assessments. The work closest to ours is~\cite{buylLargeLanguageModels2024}.
The authors analyze LLM-generated descriptions of public figures, using LLMs as evaluators, a strategy that may compound biases. 
We take the opposite approach and leverage the LLM-based sentiment classification biases to elicit biases. 
The focus on fine-grained entities and the large number of data points available enable a flexible, nuanced, and comprehensive bias analysis. 


LLMs are multilingual, and understanding linguistic and cultural variability is important for their deployment~\cite{xu2024survey}.
Existing studies reveal that LLMs often exhibit misaligned cultural and moral biases \cite{hammerlSpeakingMultipleLanguages2023, agarwalEthicalReasoningMoral2024}, with inconsistent performance and exposure biases across languages \cite{wangSeaEvalMultilingualFoundation2024}.
We contribute to this multilingual analysis by comparing LLMs prompted in six languages. 

\section{Methodology}
We systematically analyze LLMs' political bias via a task-specific evaluation framework based on target-oriented sentiment classification.
We define an inconsistency metric to quantify the TSC prediction variability. 
Our approach involves constructing a dataset of political entities and sentences, translating them into six languages, and evaluating multiple LLMs.
We diversify personal and political attributes to control confounders such as demographics and popularity. 
We also introduce a control group of synthetic entities to isolate the effects of entity-oriented biases encoded in LLMs. 
Our methodology is designed to provide statistically robust insights while remaining scalable and applicable to real-world scenarios.

\subsection{Inconsistency Metric}
We observe that LLMs inconsistently predict sentiment when changing politician names in a given sentence and leverage this variability to study biases. 
Let $\mathcal{E} = \{e_1, e_2, ..., e_n\}$ be the set of $n$ political entities, and $\mathcal{S} = \{s_1, s_2, ..., s_m\}$ the set of $m$ sentences used in experiments.
Given a sentence $s_i$, we define a sentiment prediction set:
\begin{equation}
Y_i = \{p(e_j,s_i) | e_j \in \mathcal{E}\}
\end{equation}
with: $p(e_j,s_i)$ - the predicted sentiment when replacing a placeholder in $s_i$ with entity $e_j$. 

With these notations, we define the inconsistency metric ($IC$) for an LLM model--language pair as:
\begin{equation}
IC = \frac{1}{m} \sum_{i=1}^{m} H(Y_i)
\label{eq_consistency}
\end{equation}

with: $H(Y_i)$ - the entropy of the sentiment label distribution in $Y_i$.

An unbiased LLM, a desirable behavior, would provide the same answer for a given sentence, regardless of the target entity.
The entropy value would be zero, leading to $IC=0$.
Figure~\ref{fig_teaser} empirically confirms that inconsistencies appear for all tested model-language pairs, and Equation~\ref{eq_consistency} is usable to study bias. 

\subsection{Model and Language Selection}

\paragraph{Model Selection.} 
We analyze political bias with multiple LLMs whose origin, size, and openness degree vary, including \mistral, \qwensmall, \qwenbig, \llamasmall, \llamabig, \aya, and \gptmini. 
The diversified geopolitical origin allows us to examine differences in training data composition and modeling choices. 
Varying the size of models from the same family enables the analysis of scale effects. 
Model specifications are provided in Table~\ref{tab:model_cards} in the Appendix.

\paragraph{Language Selection.} We evaluate \textit{English (eng), French (fra), Spanish (spa), Russian (rus), Arabic (ara), and Chinese (zho)} models to ensure linguistic, geopolitical, and script typology diversity. This selection spans Indo-European, Semitic, and Sino-Tibetan language families. All selected languages rank among the ten most spoken worldwide. 
Along with model diversity, the language selection ensures the representativity of the proposed political bias analysis.   

\subsection{Data Collection}

\paragraph{Entity Selection and Characterization.} 
We combine structured data from Wikidata~\cite{pellissier2016freebase} and ParlGov~\cite{doring2012parliament} to select politicians and obtain their political alignments. 
We initially retain all Wikidata entries listing "politician" among their occupations and select their political party, birth year, gender, and country-related information. 
Since political parties are sometimes associated with multiple alignments, we resort to alignment averaging to group parties into eight classical categories~\cite{heywood2021political} (see definition in section ~\ref{sec_align} of the Appendix): Far Left (FL), Left (LL), Center Left (CL), Center (CC), Center Right (CR), Right (RR), Far Right (FR), and Big Tent (BT).
Additionally, we use ParlGov political mappings (conservative vs. progressive and authoritarian vs. libertarian dimensions) to build a political compass.
We perform entity recognition using FLAIR~\cite{akbik2019flair} and entity-linking using mGENRE~\cite{de-cao-etal-2022-multilingual} over CC-News~\cite{mackenzie2020cc} to count politician mentions in the news and retain the most frequently mentioned politicians.
We then employ a hierarchical sampling algorithm to obtain a list of 1319 politicians diversified across countries and political alignments.
We detail the politician selection steps in section ~\ref{sec:balancing} of the Appendix.
Finally, a control counterpart is generated per entity by synthesizing a fake name using GPT-4, conditioned on non-political attributes (gender, birth year, and country of origin).

\paragraph{Sentence Selection and Processing.} 
We sample diversified sentences from MAD-TSC~\cite{dufraisse-etal-2023-mad}, a TSC dataset sourced from numerous European news outlets. 
These journalistic sentences often convey sentiment in a complex manner, making TSC non-trivial.
Sentences include a target entity and associated ground truth sentiment label (neutral, positive, or negative).  
Since we perform entity replacement, we design a decision diagram (see section ~\ref{sec:sentence_selection} of the Appendix) to select only sentences for which entity replacement does not produce any counterfactual or grammatically incorrect constructions.
For instance, substituting ``Donald Trump'' with another politician in a sentence containing ``US President Donald Trump'' could yield misleading implications. 
We manually create two versions of each sentence—one using a male form and one using a female form (e.g., "...in her speech" and "...in his speech") to handle gender variation. 
This process results in a set of 450 sentences comprising 150 neutral, positive, and negative examples, each available in male and female variants.
We enable multilingual analysis using professional sentence translations from MAD-TSC in French and Spanish.
We combine automatic DeepL translations with manual Russian, Chinese, and Arabic verification. 
Native speakers with advanced English skills check that the original meaning is preserved and no translation or gender-related errors occur. 

\subsection{Experimental Setup and Analysis.}

\paragraph{Prompt Design.}  We evaluate several prompting strategies to classify sentiment toward politicians using the sentences sampled from MAD-TSC. 
In an initial experiment, we test both zero-shot and few-shot prompting methods and compare constrained versus unconstrained response formats. 
We set the temperature at zero in all cases to obtain the most probable response. 
These experiments are conducted in multiple languages to ensure cross-linguistic validity and isolate political bias from artifacts introduced by prompt design. 
We summarize the obtained results in Table \ref{tab:prompt_results} and select the prompt that achieves the highest overall performance across languages and minimizes the incidence of invalid responses. 
Prompt formulations and examples are detailed in Appendix~\ref{sec:prompt_experiments}. 

\paragraph{Experimental Setup.} We construct our dataset by pairing the 450 preprocessed sentences with the 1,319 entities, resulting in 593,550 prompts per language--model combination.
With nearly 25 million data points, this setup ensures strong statistical support for the observed patterns and captures a wide range of interactions between linguistic constructs and political attributes, thereby providing a comprehensive foundation for subsequent analysis.
The TSC-based framework enables a flexible aggregation of fine-grained individual data points highlighting various bias aspects.
We present results for (1) sentiment distribution for political alignments on the left-to-right spectrum, (2) politician-level similarities, (3) 2-dimensional political compass analysis, (4) cross-language sentiment patterns, and (5) the effect of model size on bias intensity.




\section{Results} \label{sec-results}

\begin{figure*}[t]
    \centering
    \includegraphics[width=0.99\linewidth]{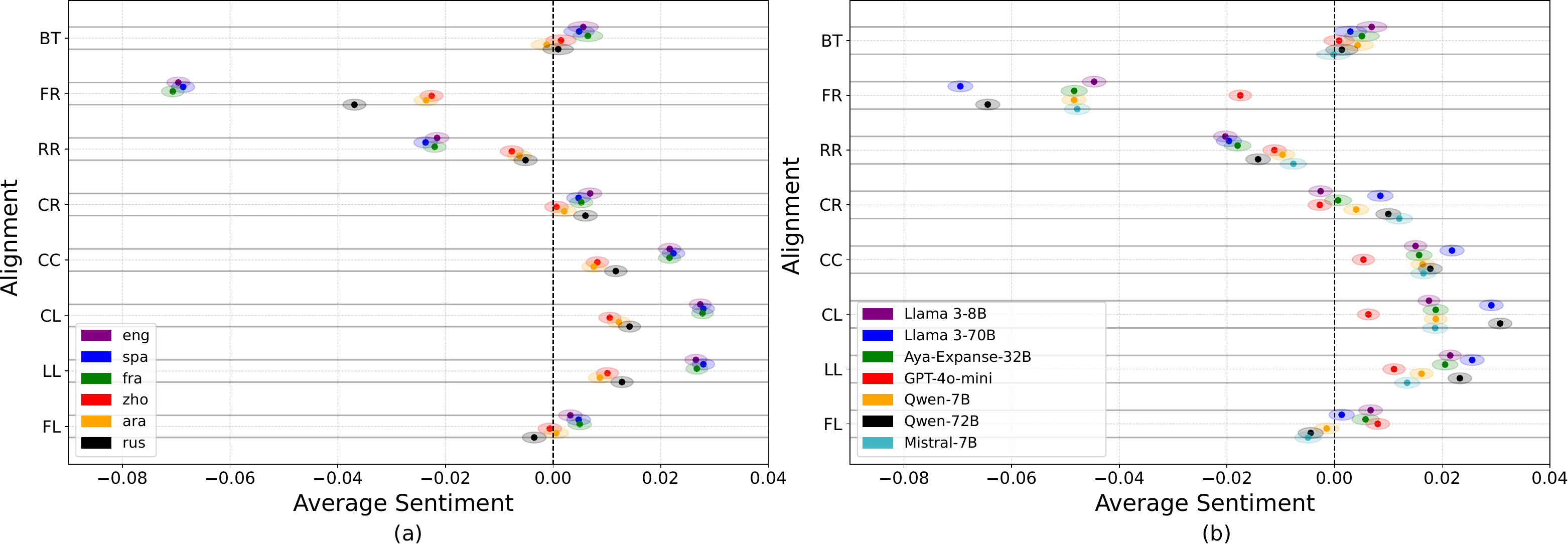}
    \vspace{-2mm}
    \caption{Average sentiment scores for languages (aggregated across all models) and for models (aggregated across all languages) per alignment. For each language or model, the averages are centered around the mean of the sentiments.   Shaded areas represent 95\% confidence intervals. The results indicate a consistent positive bias for CC, CL, and LL politicians, and a negative bias for RR and FR politicians across all languages and models. English, French, and Spanish exhibit stronger biases than Arabic, Chinese, and Russian. Additionally, larger models tend to demonstrate higher biases than smaller models.}
    \label{lang_models}
    \vspace{-3mm}
\end{figure*}

\paragraph{\textit{Sentiment analysis across models and languages reveals systematic biases based on political alignment.}} 
Figure \ref{lang_models} shows that models consistently favor left, center-left, and centrist politicians while assigning more negative sentiment to right and far-right politicians. 
These results are aligned with previous findings~\cite{gover2023political}.
Interestingly, while biases occur in all languages, they are more pronounced in English, Spanish, and French than in Russian, Arabic, and Chinese. 
A statistical test confirms the statistical significance in all languages (see section ~\ref{sec_stat_tests} of the Appendix). 
The consistency of these trends across models suggests that they are not random but may be caused by LLM training data biases. 
In the following paragraphs, we delve deeper into the influence of individual languages and models. 

\begin{figure}[t]
    \centering
    \includegraphics[width=0.99\columnwidth]{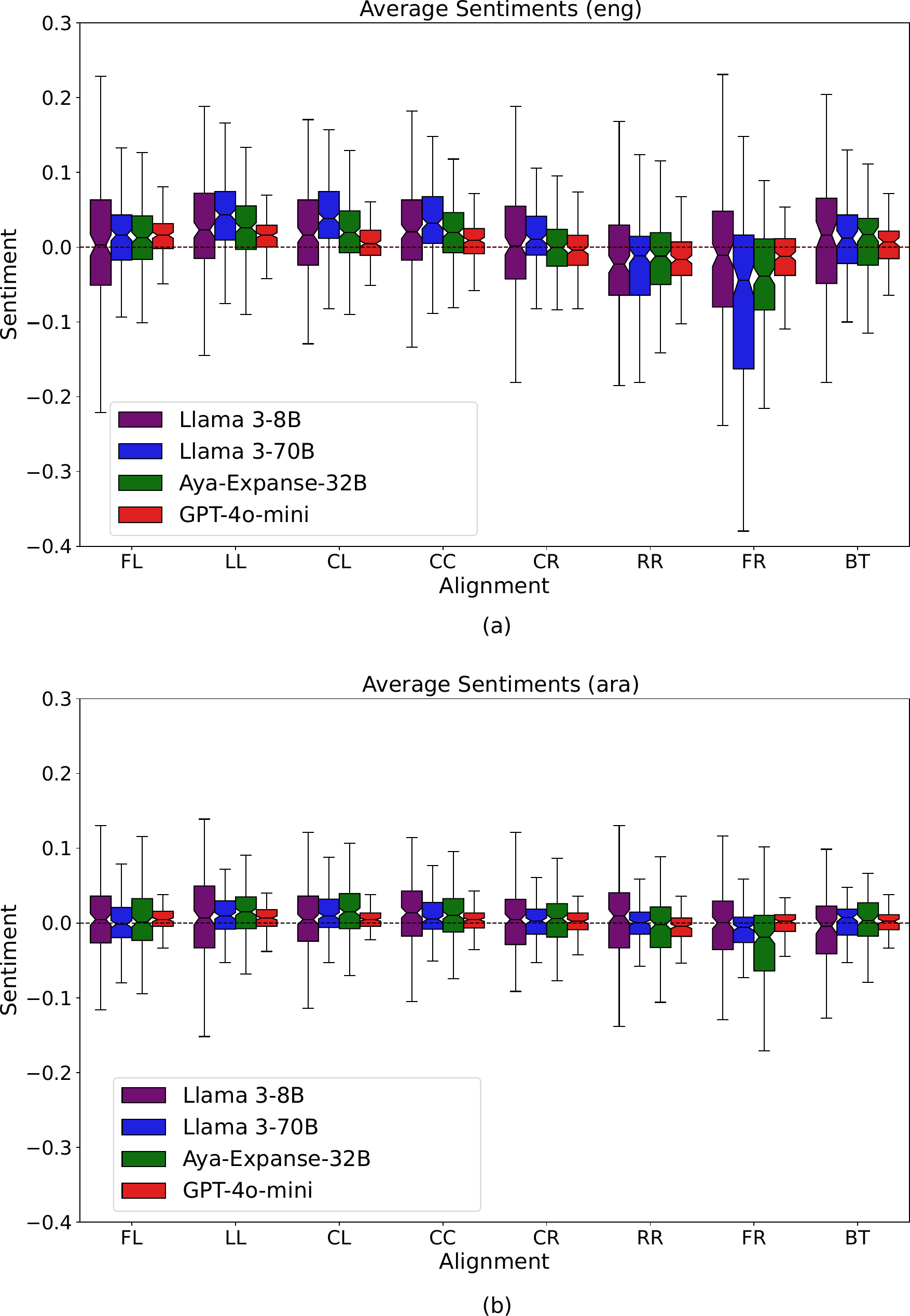}
    \caption{Boxplots depicting average sentiment scores for entities across political alignments in English and Arabic. The results reveal a pronounced bias in English, particularly a strong negative bias for far-right figures and a positive one for left figures. In contrast, biases in Arabic are less discernible, except for \aya, a model trained for multilingual tasks, which exhibits more apparent biases in Arabic as well—showing positive sentiment toward LL and CL (center-left) figures and negative sentiment toward FR figures.}
    \vspace{-3mm}
    \label{boxplot}
\end{figure}

\begin{figure*}[ht]
    \centering
\includegraphics[width=0.8\linewidth]{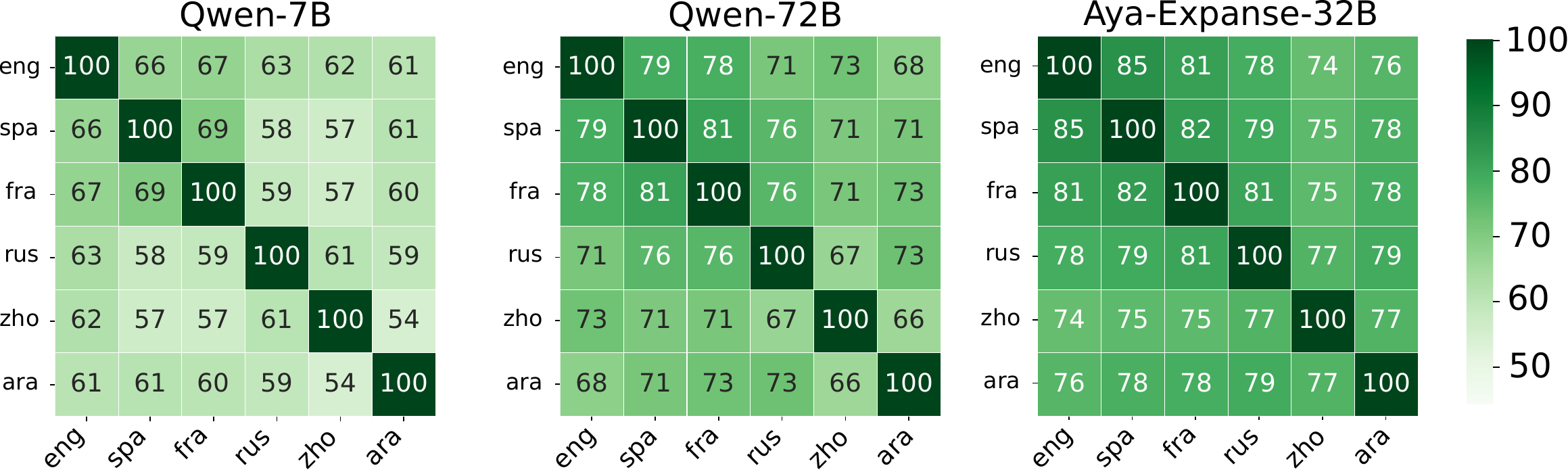}
    \caption{Jaccard similarity index between the sentiment predictions in the tested languages obtained with \qwensmall, and \qwenbig~, and \aya.}
    \label{fig_lang_corr}
    \vspace{-3mm}
\end{figure*}

\paragraph{\textit{Larger models exhibit stronger biases that are more consistent across languages.}} 
Figure \ref{lang_models} shows larger language models display more pronounced biases than their smaller counterparts. 
\qwenbig~ and \llamabig~ assign more negative sentiment to far-right and right-leaning politicians while showing stronger positive sentiment for center-right, center-left, and left-leaning politicians than \qwensmall~ and \llamasmall. 
The training corpora are unavailable, but we can safely assume they are similar within the same family of models. 
These differences are noticeable across Western and non-Western languages when analyzing individual model-language combinations in Figure~\ref{boxplot}. 
\llamasmall~ in Arabic does not exhibit a significant negative bias toward far-right politicians, the most consistent bias observed globally. 
In contrast, \llamabig~ exhibits a negative bias toward far-right figures. 
In English, biases also become more pronounced in the larger model, with stronger deviations in sentiment scores across political alignments. 
In addition to bias intensity, we study its variation across languages in Figure~\ref{fig_lang_corr} (a) by computing the correlations between sentiment predictions \qwensmall~ and \qwenbig. 
The results are obtained by averaging the Jaccard similarity indexes of individual entities for each model--language pair. 
They show that more similar TSC predictions are obtained with larger models. 
These results suggest that as models scale up, their positive and negative biases tend to intensify but also become more consistent across languages. 
This is possibly due to larger LLMs' increased capacity to internalize implicit patterns from training data.

\paragraph{\textit{Aya-Expanse-32B exhibits stronger biases in non-Western languages.}} Figure \ref{boxplot} (b) shows \aya~ displays higher biases for Arabic. 
While the figure specifically highlights Arabic, this pattern holds consistently for Russian and Chinese.
This is the only model specifically trained for multilingual tasks in our set of models. 
An explanation comes from Aya's training methodology, which relies heavily on translating data from high-resource languages (e.g., English) to others. 
Figure \ref{fig_lang_corr} also highlights that Aya's predictions are more similar across languages than those of Qwen models. 
While Aya's multilingual training is effective for cross-lingual generalization, it may also lead to a more uniform propagation of biases across diverse linguistic contexts. 
These findings underscore the trade-offs in multilingual model design and the need to consider how training data and methodologies influence bias amplification carefully.

\begin{figure}[t]
    \centering
    \includegraphics[width=0.99\columnwidth]{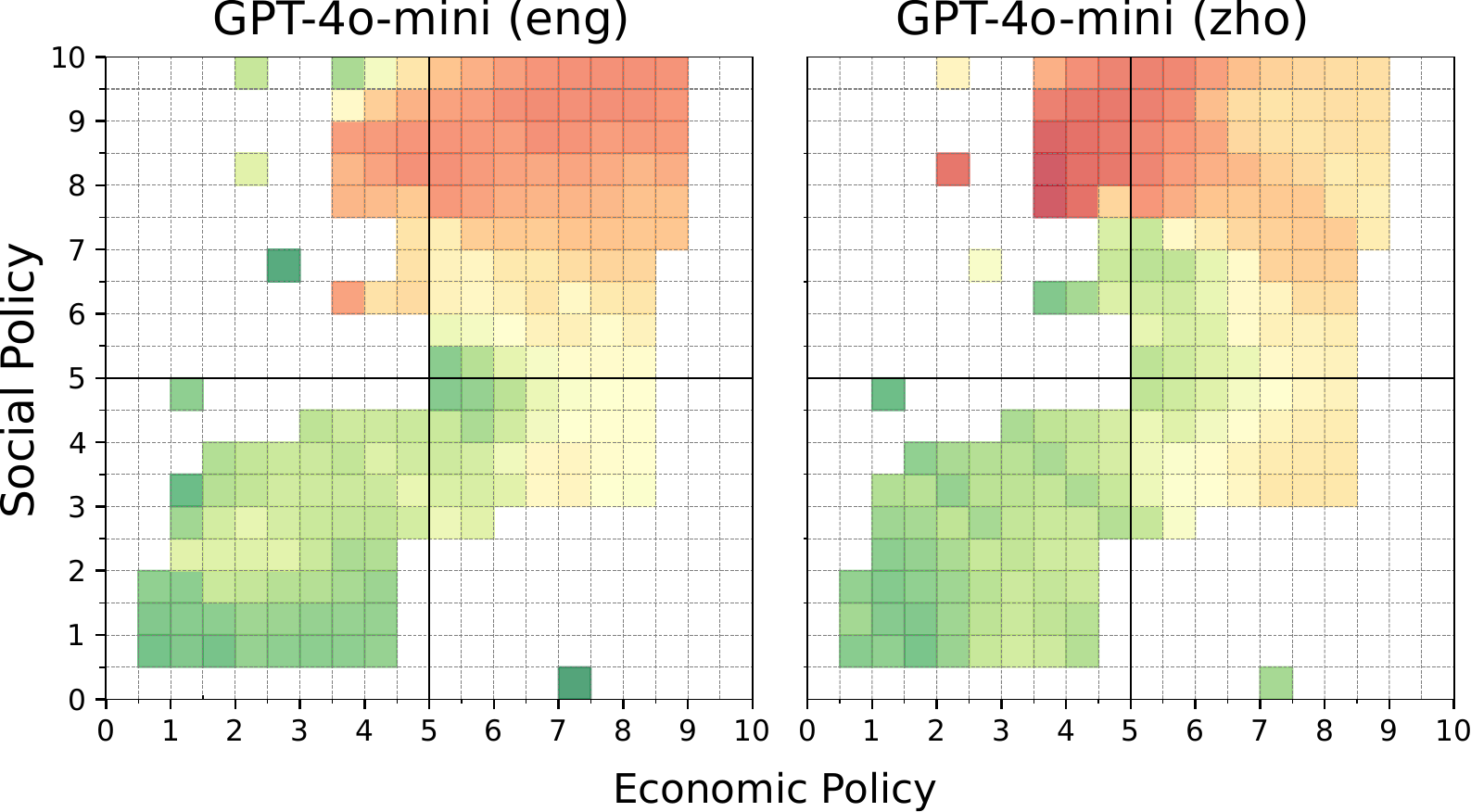}
    \caption{Political compasses showing sentiment bias for GPT-4o-mini in English (left) and Chinese (right). The y-axis represents the social policy spectrum (0: socially progressive, 10: socially conservative), and the x-axis represents the economic policy spectrum (0: fiscally progressive, 10: fiscally conservative). Parties are positioned using ParlGov data, with colors indicating the average sentiment of affiliated politicians (red: negative, green: positive). Blank squares denote no corresponding party. Left-libertarian parties consistently receive positive sentiment, while right-authoritarian parties show negative sentiment, highlighting consistent ideological biases across languages.}
    \vspace{-3mm}
    \label{compass}
\end{figure}

\paragraph{\textit{LLMs have a left-libertarian bias.}} Using ParlGov data, which assigns parties a progressive-conservative and authoritarian-libertarian score, we map politicians onto a 10×10 political compass grid. We compute the average sentiment score of all associated politicians for each square, then average the results over neighboring squares to highlight broader trends.
As shown in Figure \ref{compass}, the sentiment distribution for GPT-4o-mini in English and Chinese aligns with findings from previous studies~\cite{hartmannPoliticalIdeologyConversational2023,motokiMoreHumanHuman2024}, reinforcing the conclusion that ChatGPT exhibits a left-libertarian bias. 
Sentiment analysis across the political compass reveals a consistent pattern: politicians associated with left-libertarian positions receive more positive sentiment, while those aligned with right-authoritarian positions receive more negative sentiment. Additionally, our approach enables a more fine-grained analysis, revealing that centrist parties also tend to receive higher sentiment scores. This trend holds across all languages and models, further supporting the robustness of these patterns.

\begin{figure}[t]
    \centering
    \includegraphics[width=0.99\columnwidth]{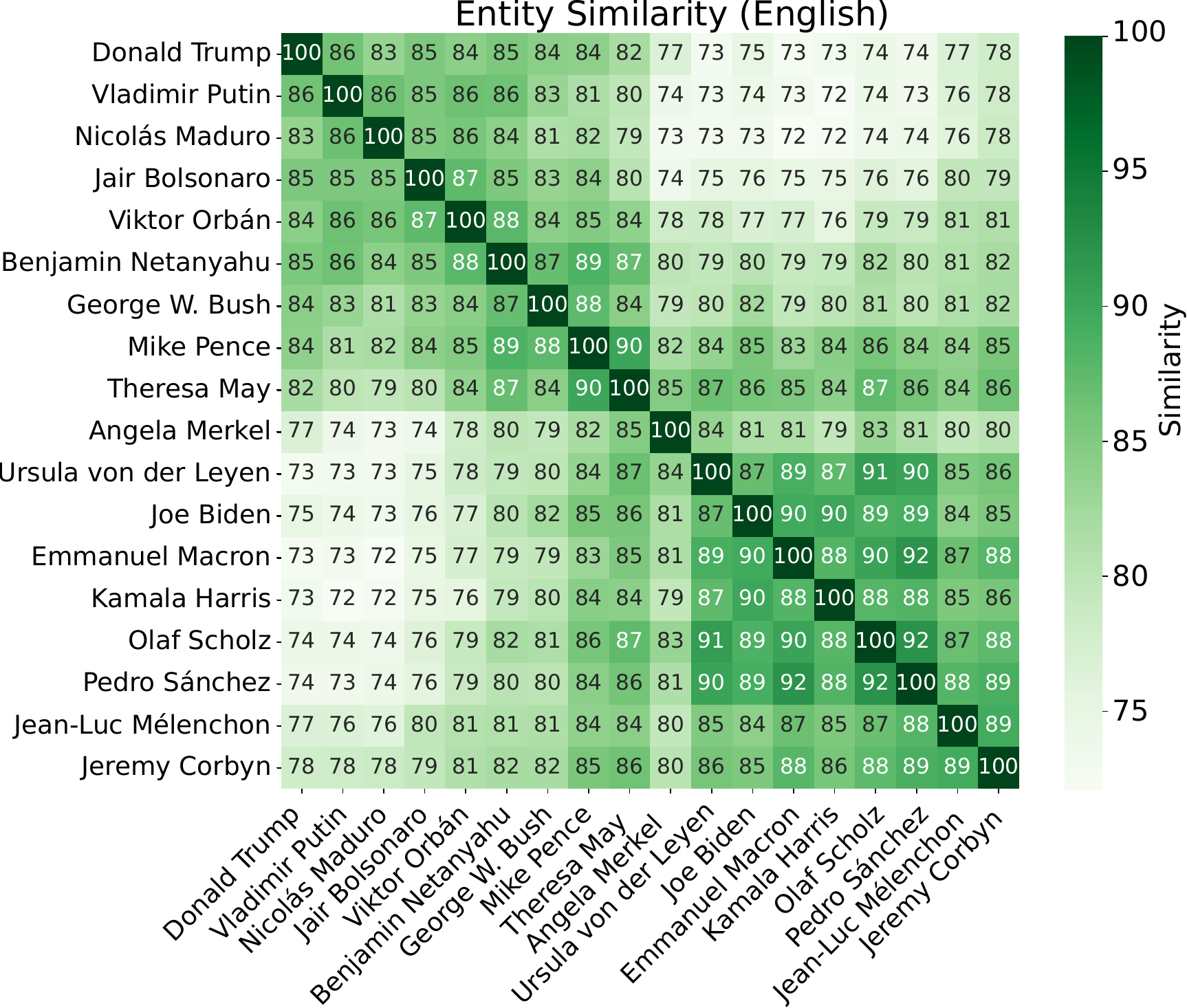}
    \caption{Similarity matrix of entities in English, where each cell represents the similarity between two entities (scaled to 100). Greener squares indicate higher similarity. The matrix reveals two distinct blocks: one in the top left and another in the bottom right, suggesting a divide between entities potentially aligned with authoritarian tendencies and those associated with liberal Western democracies. Closer similarity values reflect aligned sentiment patterns, potentially indicating ideological or contextual affinities.}
    \vspace{-3mm}
    \label{similarity_matrix}
\end{figure}

\paragraph{\textit{LLMs have an internal representation of entities.}}
To test this hypothesis in sentiment analysis tasks, we construct a sentiment matrix $M_k$ for each entity $e_k$, where the element $(i, j)$ of the matrix corresponds to the output sentiment from model $j$ when given input sentence $i$ with target entity $e_k$. We then compute the cosine similarity between matrices \( M_{k_1} \) and \( M_{k_2} \). For each column \( j \), we calculate the cosine similarity between the corresponding column vectors of \( M_{k_1} \) and \( M_{k_2} \):

\[
K_j(M_{k_1}, M_{k_2}) = \frac{\langle M_{k_1}^{(j)}, M_{k_2}^{(j)} \rangle}{\|M_{k_1}^{(j)}\| \cdot \|M_{k_2}^{(j)}\|},
\]

where \( M_{k_1}^{(j)} \) and \( M_{k_2}^{(j)} \) denote the \( j \)-th column vectors of \( M_{k_1} \) and \( M_{k_2} \), respectively, \( \langle \cdot, \cdot \rangle \) is the dot product, and \( \|\cdot\| \) is the Euclidean norm. The final cosine similarity is the average of these values across all columns:

\[
K(M_{k_1}, M_{k_2}) = \frac{1}{J} \sum_{j=1}^J K_j(M_{k_1}, M_{k_2}),
\]

where \( J \) is the total number of columns. 

Using this similarity measure, we construct a similarity matrix (Figure~\ref{similarity_matrix}), where each cell represents the similarity between two entities. The matrix reveals two distinct blocks: one in the top left and another in the bottom right. This division suggests a separation between entities associated with liberal democracies and those potentially aligned with authoritarian tendencies. Entities closer in similarity exhibit more aligned sentiment patterns, reflecting ideological or contextual affinities.
For example, \textit{Pedro Sánchez} and \textit{Olaf Scholz}, both European socialist leaders, show high similarity (\textcolor{darkgreen}{+0.92}), likely due to shared political ideology and regional context. Similarly, \textit{Joe Biden} and \textit{Kamala Harris}, who share a political affiliation and roles within the same administration, are closely aligned (\textcolor{darkgreen}{+0.9}). \textit{Jean-Luc Mélenchon} and \textit{Jeremy Corbyn} (\textcolor{darkgreen}{+0.89}), both left-wing politicians advocating for progressive policies, also exhibit significant similarity. These patterns emerge in all the tested languages and suggest that LLMs encode meaningful representations of entities, capturing real-world ideological and contextual distinctions even in sentiment analysis tasks.

\begin{figure*}[t]
    \centering
    \includegraphics[width=0.99\textwidth]{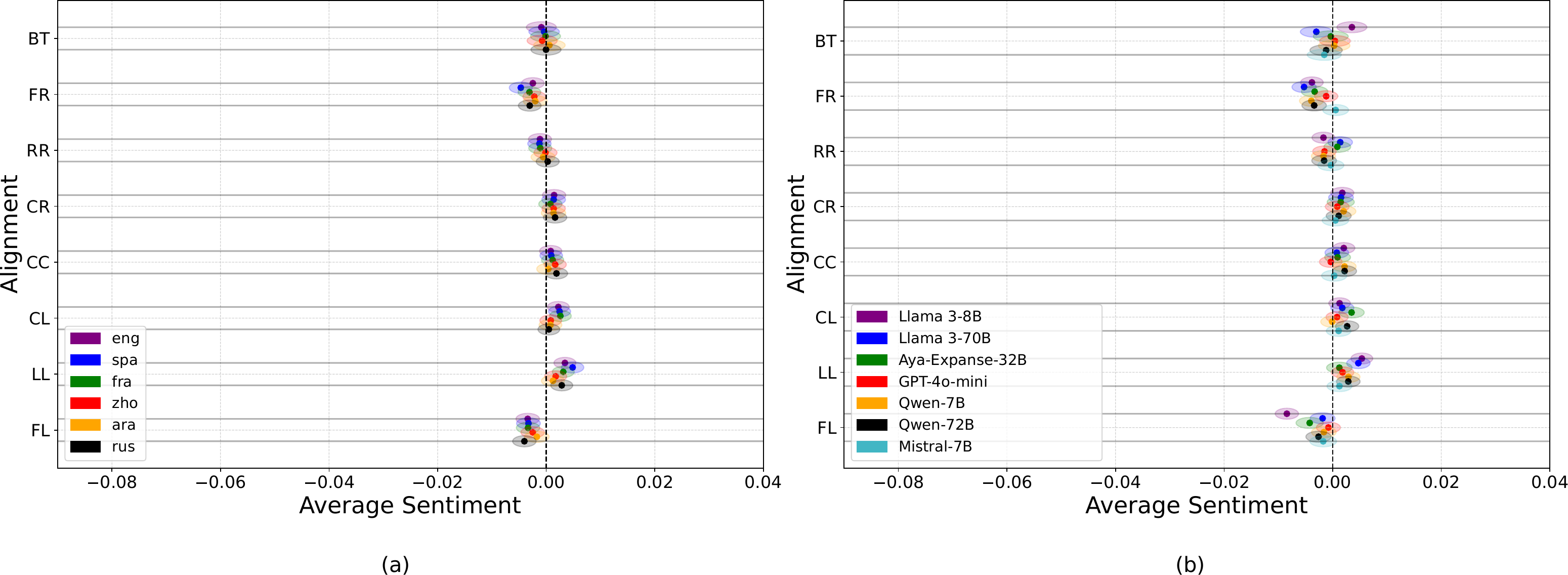}
    \vspace{-2mm}
    \caption{Average sentiment scores for languages and for models using a set of fake names reflecting non-political attributes (gender, birth year, and country of origin) of our original entities. Shaded areas represent 95\% confidence intervals. The results show that the biases for political entities largely disappear, although some anomalies persist. These anomalies are attributed to LLMs rating female names higher (often corresponding to left-leaning entities) and Russian or non-Western names lower (often corresponding to far-left or far-right leaning entities). This suggests that TSC biases are primarily driven by political associations rather than non-political attributes.}
    \label{lang_models_np}
    \vspace{-3mm}
\end{figure*}

\paragraph{Control Group and Mitigation.} 
We detail the control group experiment summarized in Figure~\ref{fig_teaser} whose objective is two-fold: (1) it confirms that political attributes are the primary source of bias by isolating the influence of political associations on model behavior, and (2) it provides a method to mitigate these biases, making sentiment analysis fairer and more accurate. 
For each entity, we use an LLM to generate a fake name that preserves non-political attributes (gender, birth year, and country of origin). 
Running simulations with the control group reveals significant improvements in model consistency and slight gains in accuracy, as shown in Figure~\ref{fig_teaser}. This suggests that removing political attributes reduces inconsistencies in model responses across entities.

Figure~\ref{lang_models_np} demonstrates that political biases are significantly reduced compared to the original results in Figure~\ref{lang_models}. 
However, some differences persist and are explained by demographic factors.
First, female names (\textcolor{darkgreen}{+0.03}), which are overrepresented in the left alignment, have more positive predictions than their male counterparts (\textcolor{burgundy}{-0.01}). 
Second, LLMs tend to rate Russian and non-Western names lower (often linked to far-left or far-right entities). These residual biases suggest that while political attributes are the primary driver of bias, non-political attributes still exert a minor influence. 

This control group approach confirms that political attributes are the primary source of bias and provides a practical method to mitigate them in TSC. By isolating and addressing these sources, we can enhance the LLM fairness and reliability.

\section{Conclusion}

Understanding political bias in LLMs is crucial as their real-world use grows. 
In this study, we present a novel approach using target-oriented sentiment inconsistencies to quantify these biases. 
We systematically varied political entities in controlled sentence templates to obtain a massive TSC prediction dataset and aggregate them to analyze different bias facets.
Our results have several important implications.
First, we contribute to studying the LLMs' political biases by introducing a flexible framework that builds on TSC, a fine-grained NLP task. 
This modeling enables a flexible aggregation of results, making our approach complementary to existing ones. 
We emphasize the roles of languages and LLM sizes and analyze political alignments and individual entities.  
Our results highlight systematic political biases in LLMs, with more positive sentiment assigned to left-leaning politicians and more negative sentiment to far-right figures. 
Bias intensity is stronger in English, French, and Spanish but exists across all tested languages. 
We find that larger models accentuate bias and increase similarities between predictions in different languages. 
These findings support previous work~\cite{yin2022geomlama} showing that native language is not the best choice for probing regional knowledge when alternatives exist.
We also highlight similarities between the LLMs representations of individual politicians with similar alignments. 

Second, the prediction inconsistencies significantly affect the LLM use for target-oriented sentiment classification. 
This finding, added to the fact that small models are still competitive in TSC~\cite{bucher2024fine}, questions the LLM robustness for this particular task despite their easy deployment. 
More generally, the observed inconsistency leads us to recommend caution when employing LLM for bias-prone tasks, particularly those involving subjectivity. 

Third, these biases can be mitigated to a certain extent. 
We tested a post-training strategy by replacing politician names with more neutral alternatives.
This strategy reduces sentiment inconsistencies and also brings small accuracy gains. 
However, addressing bias in one task does not guarantee neutrality in others. 
Future work should explore how bias manifests in different applications and across a broader range of entities.
Equally important, LLM creators could use our findings to integrate bias mitigation in subjective tasks during training.     

Finally, the proposed strategy is model-agnostic and adaptable to other languages. 
We will release the code and data to contribute to systematic and reproducible political bias evaluation. 

\section{Limitations}

\paragraph{Analysis Scope.} 
Our study complements those focusing on controlled text generation~\cite{gover2023political,bangMeasuringPoliticalBias2024}.
We confirm the existence of political bias in LLMs through the TSC lens, an important downstream task.
The study could be further explored in two directions for greater comprehensiveness.
First, it would be interesting to test whether our conclusions hold for other downstream tasks having a subjective component, including stance or hate speech detection.
Second, we could explore LLM biases for entities other than politicians to broaden the scope.
We could test other entities within the political domain, such as political parties; and non political ones, including public organizations, companies, or products. 
Such extensions depend on the availability of labeled datasets for the cited tasks and entity types.

\paragraph{Data Representativeness.} 
Our dataset includes a diverse set of politicians but remains dominated by figures frequently mentioned in Western media, which make up the majority of CC-News.
This modeling choice is intentional since political bias occurs for entities often encountered in the training data, as shown by the control experiment. 
As a result, some countries remain underrepresented in the entity set. 
Similarly, the sentences used in the experiments are primarily drawn from European political discourse. 
We filtered sentences focused on a narrow context to enable controlled comparisons when varying the politician names.
However, we cannot guarantee that they capture the nuances of the political debate globally. 

\paragraph{Bias Mitigation and Trade-offs.} Including control entities (non-political names) helps distinguish biases related to political attributes and effectively reduces entity-related bias in our context.
However, bias subsists to some degree, and the method could be further improved by checking that control entities' first and last names are not politically charged. 
Equally, the replacement approach also reduces some contextual information. 
While necessary for isolating the effects of political affiliation, this trade-off may limit insights into how biases manifest in usual political discussions in other contexts.
More advanced political bias mitigation techniques would require in-depth model adaptations beyond this contribution's scope. 

\paragraph{Evolving Nature of Biases.} 
The biases observed reflect the state of political text included in the LLMs' training dataset. 
For instance, the sentiment expressed about a politician's or a political alignment's actions can vary drastically over time. 
The reported findings will probably evolve as models are updated with new data and political landscapes shift, and 
future research should periodically reassess bias patterns to account for these changes.

\paragraph{Model and Language Representativeness.} 
We experimented with seven models and six languages representative of current LLMs, enabling meaningful comparisons across these two axes.
The results highlight bias for all model-language combinations, with some variability for specific combinations, allowing us to present robust findings. 
The study could be further generalized by including more models and languages, but such an extension is challenging.
First, we performed initial tests with more open-source models but discarded them due to low accuracy and/or the impossibility of obtaining usable inferences. 
Second, we tested many entity-sentence combinations, and the cost of running complete experiments with multiple models, particularly in a few-shot learning setting, exceeded our financial means.  
We selected six widely used languages from linguistic groups representing different geopolitical alignments and found interesting differences between models in these languages.  
Integrating other languages would further improve the results' robustness.
It would be particularly interesting to test biases for low-resourced languages. 

\section{Ethical Considerations}
This study investigates political bias in large language models (LLMs) to improve fairness and transparency in AI systems. We prioritize transparency by releasing our methodology, data, and code for reproducibility, and focus on systemic bias patterns rather than individuals to avoid reinforcing stereotypes or misinformation. Our analysis includes diverse politicians across regions, ideologies, and languages to ensure balanced representation. We acknowledge that findings are influenced by dataset and model choices, and caution against misusing results to generalize about political groups or imply intentional bias in LLMs. This work does not endorse political positions and aims to support safer, more equitable AI systems for politically sensitive applications.

\section*{Acknowledgements}
This work has been funded by the BOOM ANR Project - ANR-20-CE23-0024 and the BPI funded project OpenLLM-France.
It was made possible by the use of the FactoryIA supercomputer, financially supported by the Ile-de-France Regional Council.

\bibliography{anthology,custom}

\appendix

\section{Alignment Computation for Political Entities}
\label{sec:alignments}

To define political alignment for entities, we link them to their respective political parties using Wikidata. The political alignment of each entity is then inferred from the alignment information associated with their party in Wikidata. However, this approach introduces two key challenges:

\begin{enumerate}
    \item \textbf{Multiple Party Associations:} Entities may be associated with more than one political party.
    \item \textbf{Multiple Alignment Associations:} Political parties may be associated with more than one political alignment.
\end{enumerate}

\subsection{Resolving Multiple Party Associations}

Entities can have multiple party affiliations over time. For example, Donald Trump is associated with several parties, including the Republican Party, Reform Party of the United States of America, Democratic Party, and independent politician. Wikidata ranks these associations into three categories: \textit{Preferred Rank}, \textit{Normal Rank}, and \textit{Deprecated Rank}. To resolve this issue, we apply the following procedure:

\begin{itemize}
    \item First, select the entry with the \textit{Preferred Rank}, provided it has a non-blank name.
    \item If no such entry exists, iterate through the remaining entries, ordered by their \textit{end time} value. Select the party with either no end time or the most recent end time.
\end{itemize}

This method ensures that each politician is associated with the most accurate and relevant political party. We further validate this by manually verifying the results.

\paragraph{Resolving Multiple Alignment Associations}

Political parties may also have multiple alignment associations. For instance, the Renaissance Party of France is associated with three alignments: \textit{centrism}, \textit{centre-left}, and \textit{centre-right}. To address this, we observe the distribution of alignments across all entities and note that centrist alignments are overrepresented compared to far-left or far-right alignments. To ensure balanced representation and assign a single alignment to each party, we employ the \textit{Alignment Computation} algorithm (see Algorithm~\ref{alg:process_alignment}).

The algorithm works by averaging the alignments and then rounding them away from the center. For example, the Renaissance Party's alignments (\textit{centre-left}, \textit{centrism}, and \textit{centre-right}) are averaged and rounded to \textit{CC} (center). This approach ensures a balanced and consistent representation of political alignments across the dataset.

\begin{algorithm}
\caption{Alignment Computation}
\label{alg:process_alignment}
\textbf{Input:} \quad alignments, a list of alignments. \\
\textbf{Output:} \quad A single alignment label computed from the inputs.
\begin{algorithmic}[1]
    \State Define \texttt{mapping} as:
    \begin{itemize}
        \item "Far-Left" $\to$ -3
        \item "Left-Wing" $\to$ -2
        \item "Center-Left" $\to$ -1
        \item "Center" $\to$ 0
        \item "Center-Right" $\to$ 1
        \item "Right-Wing" $\to$ 2
        \item "Far-Right" $\to$ 3
    \end{itemize}
    \If{length(\texttt{alignments}) = 1}
        \State \Return \texttt{alignments[0]}
    \EndIf
    \medskip
    \If{"Big Tent" is in \texttt{alignments}}
        \State Remove "Big Tent" from \texttt{alignments}
        \If{\texttt{alignments} is empty}
            \State \Return "Big Tent"
        \EndIf
    \EndIf
    \medskip
    \State \texttt{sum} $\gets$ 0
    \For{each \texttt{alignment} in \texttt{alignments}}
        \State \texttt{sum} $\gets$ \texttt{sum} + \texttt{mapping[alignment]}
    \EndFor
    \State \texttt{avg} $\gets$ \texttt{sum} / length(\texttt{alignments})
    \medskip
    \State \texttt{rounded} $\gets$ \Call{RoundAwayFromZero}{\texttt{avg}}
    \medskip
    \For{each (\texttt{label}, \texttt{score}) in \texttt{mapping}}
        \If{\texttt{score} = \texttt{rounded}}
            \State \Return \texttt{label}
        \EndIf
    \EndFor
    \State \Return \texttt{None}
\end{algorithmic}
\end{algorithm}

\section{Balancing the Dataset} \label{sec:balancing}

After associating each entity with a political alignment, we address the need to balance the dataset across both political alignments and countries of origin. This ensures that no bias is introduced by the over- or underrepresentation of specific countries, promotes diversity among entities, and guarantees that aggregated results over political alignments are statistically significant.

\begin{figure}[t]
    \centering
    \includegraphics[width=0.99\columnwidth]{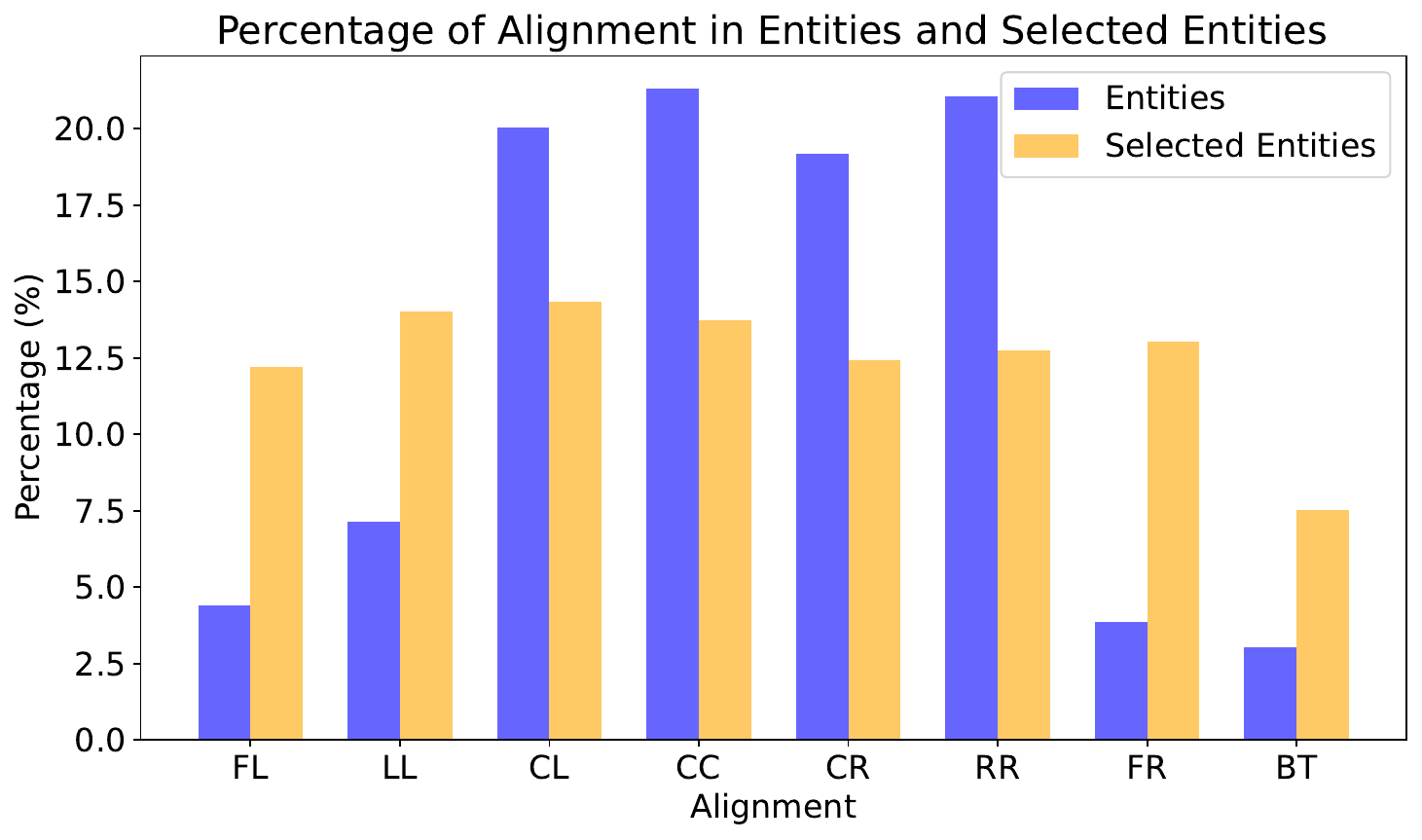}
    \caption{Alignment distribution among entities extracted from the CC-News corpus (in blue) and among entities selected and used for the experiment (in orange). While the original corpus shows an uneven distribution of political alignments, our selection process successfully balanced the representation of different alignments for the experiment.}
    \label{fig:entities_distribution}
\end{figure}

\begin{figure*}
    \centering
    \includegraphics[width=0.99\linewidth]{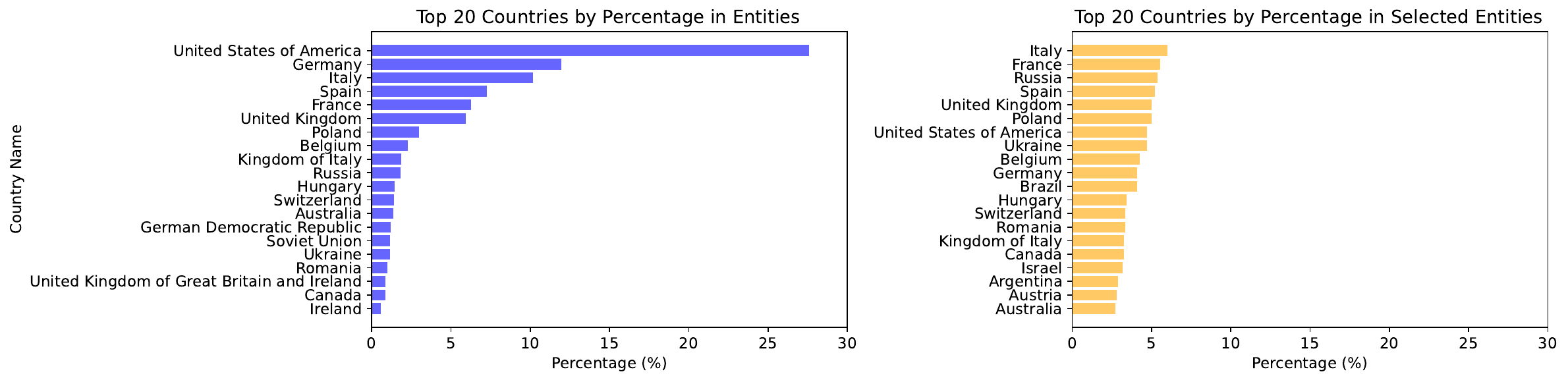}
    \caption{Countries distribution among entities extracted from the CC-News corpus (in blue) and among entities selected and used for the experiment (in orange).}
    \label{fig:entities_countries}
\end{figure*}

In Figure~\ref{fig:entities_distribution}, the blue bars illustrate the initial distribution of entities across political alignments. We observe that center-left (CL), center (CC), center-right (CR), and right (RR) alignments are overrepresented, while other alignments are underrepresented—despite our earlier step of averaging alignments away from the center. Additionally, Figure~\ref{fig:entities_countries} reveals that entities from the United States constitute over 25\% of the dataset, indicating a significant geographic imbalance.

To address these issues, we devise a hierarchical sampling algorithm (see Algorithm~\ref{alg:hierarchical_sampling}). This algorithm samples a limited number of politicians from each country while ensuring oversampling of underrepresented alignments and undersampling of overrepresented ones. The result is a balanced dataset of 1,319 entities, as shown in Figures~\ref{fig:entities_distribution} and~\ref{fig:entities_countries}, which demonstrates successful balancing across both political alignments and countries.

One notable limitation is the inability to balance the dataset across gender. Male entities constitute 75\% of the dataset and are significantly overrepresented in certain alignments (e.g., 90\% in Far-right). Achieving gender balance while maintaining alignment and country balance proved infeasible, as it would have required excluding a substantial portion of the data or introducing artificial imbalances in other dimensions.

\begin{algorithm}
\caption{Hierarchical Sampling of Politicians}
\label{alg:hierarchical_sampling}
\textbf{Input:} \quad selected\_countries, entities \\
\textbf{Output:} \quad sampled\_politicians
\begin{algorithmic}[1]
\State sampled\_politicians $\gets$ empty list
\For{each country in selected\_countries}
    \For{each alignment in \{"FL", "BT"\}}
        \State Let \textbf{Subset\_FL\_BT} $\gets$ filter(entities, where Country = country country and Alignment = alignment)
        \State Append top $k_1$ entities from \textbf{Subset\_FL\_BT} to sampled\_politicians
    \EndFor
    \For{each alignment in \{"CL", "CR"\}}
        \State Let \textbf{Subset\_CL\_CR} $\gets$ filter(entities, where Country = country and Alignment = alignment)
        \State Append top $k_2$ entities from \textbf{Subset\_CL\_CR} to sampled\_politicians
    \EndFor
    \For{each alignment in \{"CC"\}}
        \State Let \textbf{Subset\_CC} $\gets$ filter(entities, where Country = country and Alignment = "CC")
        \State Append top $k_3$ entities from \textbf{Subset\_CC} to sampled\_politicians
    \EndFor
    \For{each alignment not in \{"FL", "BT", "CL", "CR", "CC"\}}
        \State Let \textbf{Subset\_Other} $\gets$ filter(entities, where Country = country and Alignment = alignment)
        \State Append top $k_4$ entities from \textbf{Subset\_Other} to sampled\_politicians
    \EndFor
\EndFor
\State \Return sampled\_politicians
\end{algorithmic}
\end{algorithm}

\section{Control Group Generation}
\label{sec:control_group}

\begin{figure}[t]
    \centering
    \includegraphics[width=0.99\columnwidth]{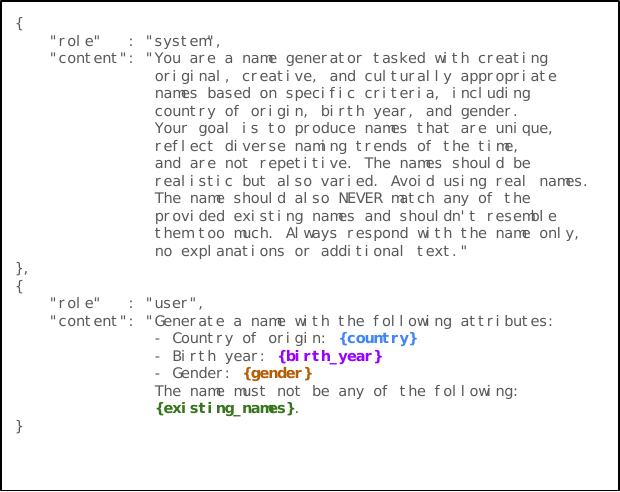}
    \caption{Prompt used for fake name generation. Placeholders \{country\}, \{birth\_year\}, and \{gender\} are replaced by the corresponding attributes of the original entity. \{existing\_names\} is constructed iteratively and consists of the names previously generated to ensure no duplicates.}
    \label{fig:prompt}
\end{figure}

To generate a control group of fictional entities, we use GPT-4 with the prompt detailed in Figure~\ref{fig:prompt}. The prompt instructs the model to create original, culturally appropriate names based on specific criteria: country of origin, birth year, and gender. The generated names must be unique and realistic, and the names must not resemble any existing names provided in the input. This ensures that the control group consists of entirely fictional entities with no overlap or similarity to real-world politicians.

The prompt is structured as follows:
\begin{itemize}
    \item The \textit{system role} defines the task as generating culturally and temporally appropriate names that are unique and avoid repetition.
    \item The \textit{user role} specifies the attributes for each name: country of origin, birth year, and gender, along with a list of existing names to avoid.
\end{itemize}

By using this approach, we ensure that the control group mirrors the distribution of gender, birth year, and country of origin present in our dataset of real entities. This alignment allows us to isolate the impact of political attributes when rerunning simulations, as any observed differences can be more confidently attributed to political factors rather than demographic or geographic biases.

This method provides a robust foundation for comparative analysis, enabling us to evaluate how political attributes influence outcomes independently of other variables.

\section{Political Alignment Definitions}
\label{sec_align}

We define the eight political alignments used in the analysis below:
We define the eight political alignments used in the analysis below:
\begin{itemize}
    \item \textbf{Far Left (FL)}: Political positions typically associated with advocacy for extensive structural changes, often including policies related to collective ownership and redistribution of resources.

    \item \textbf{Left (LL)}: Political positions generally characterized by support for social welfare systems, progressive taxation, and policies aimed at reducing economic disparities.

    \item \textbf{Center Left (CL)}: Political positions that combine support for social welfare and progressive policies with a focus on moderate and incremental reforms.

    \item \textbf{Center (CC)}: Political positions that emphasize neutrality, compromise, and policies based on pragmatic considerations rather than ideological extremes.

    \item \textbf{Center Right (CR)}: Political positions typically supporting free-market policies, limited government intervention, and traditional social structures, while allowing for some flexibility on social issues.

    \item \textbf{Right (RR)}: Political positions characterized by advocacy for free markets, limited government, national sovereignty, and the preservation of traditional institutions.

    \item \textbf{Far Right (FR)}: Political positions often associated with strong nationalism, restrictive immigration policies, and opposition to progressive social changes.

    \item \textbf{Big Tent (BT)}: Political movements or parties that aim to include a wide range of ideological perspectives, often prioritizing broad coalition-building over strict adherence to a single ideology. This category may also include movements or parties that are difficult to classify due to their eclectic or evolving policy positions, as well as those that adapt their stances to appeal to diverse constituencies.
\end{itemize}

\section{Sentence Selection}
\label{sec:sentence_selection}

To ensure that LLMs are prompted with sentences that do not contain counterfactual information or identifiable context, we implement a rigorous selection process. This process ensures that sentences are general enough to allow entity substitution without introducing inconsistencies or requiring the model to correct factual errors. The selection criteria are summarized in the decision diagram in Figure~\ref{fig:decision_diagram} and involve the following steps:

\begin{figure*}
    \centering
    \includegraphics[width=0.99\linewidth]{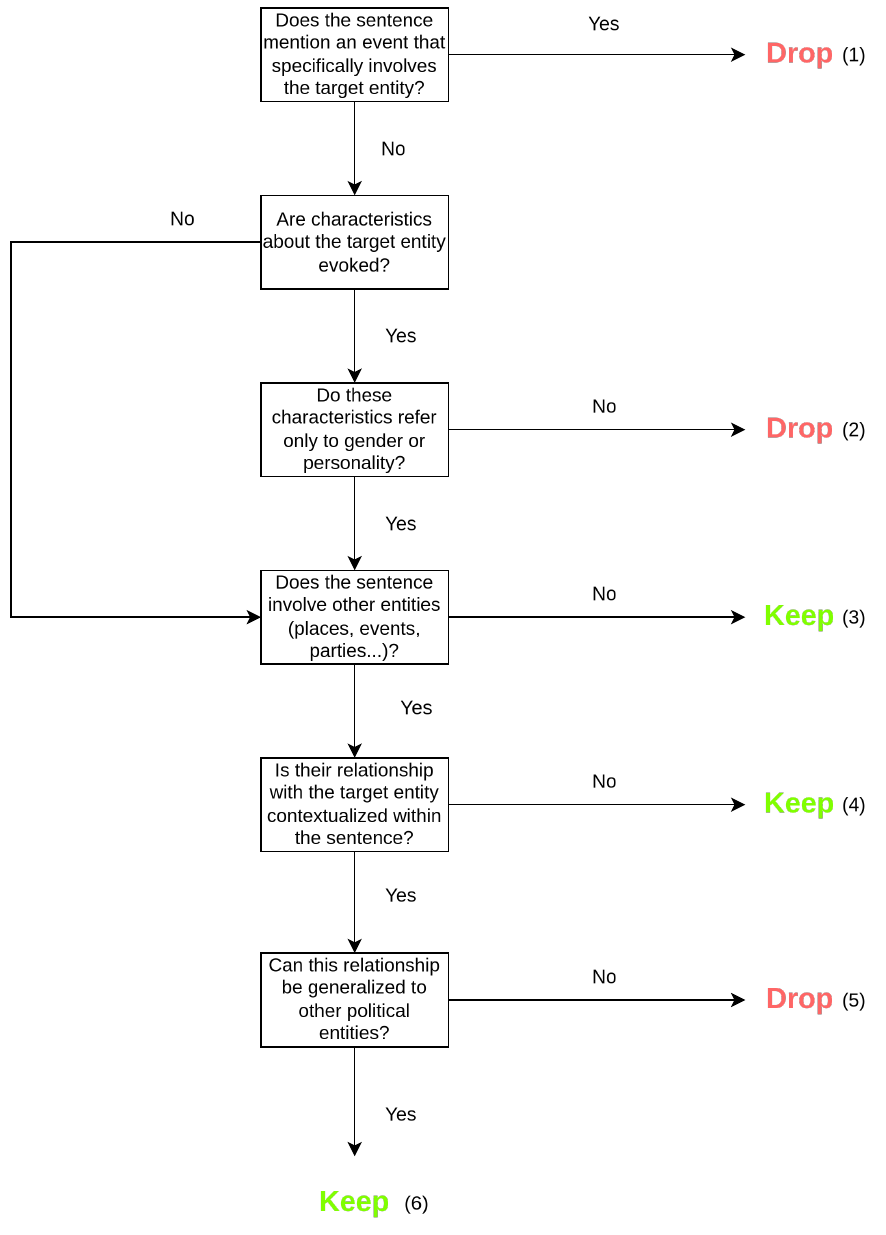}
    \caption{Sentence selection diagram. This decision diagram outlines the process of either rejecting or selecting sentences for the experiment. Sentences are rejected if switching entities introduces counterfactual information or specific unwanted contexts. However, sentences are manually adjusted for gender agreement (feminine and masculine forms), so gender-related characteristics alone are not sufficient grounds for rejection.}
    \label{fig:decision_diagram}
\end{figure*}

\begin{itemize}
    \item Check for information that may identify the original entity, such as specific events or non-generalizable characteristics (e.g., unique roles or titles). Gender and personality traits are excluded from this step, as we adjust for gender and personality traits are not overly specific.
    \item Check for additional entities in the sentence and ensure that their relationship with the main entity is not too specific to generalize.
\end{itemize}

The table \ref{tab:sentence_selection} provides examples of sentence evaluation, including the original sentence, the step applied, the decision, and the rationale behind acceptance or rejection. The original entities in each sentence are highlighted in \textbf{bold}.

\begin{table*}
\centering
\caption{Sentence Selection Process}
\label{tab:sentence_selection}
\begin{tabular}{|p{8cm}|c|c|p{4cm}|}
\hline
\textbf{Original Sentence} & \textbf{Step} & \textbf{Decision} & \textbf{Rationale} \\
\hline
``On October 4, \textbf{Angela Merkel} is to meet with leaders of the Social Democratic Party (SPD) to begin what will likely be extended and complex negotiations on a possible grand coalition.'' & 1 & \textcolor{burgundy}{Rejected} & The sentence mentions a specific event involving Angela Merkel and the SPD, which cannot be generalized to other entities, as entities from other countries cannot form a grand coalition with a german party. \\
\hline
``IMF director \textbf{Christine Lagarde} has urged Eurozone countries to 'immediately enforce the agreements of the July 21 summit, because the time factor is crucial', while US treasury secretary Tim Geithner warned that 'preventing a default of Greece is more important than sustaining European growth'.'' & 2 & \textcolor{burgundy}{Rejected} & ``IMF Director'' is a specific role that cannot be generalized to other entities. \\
\hline
``Already, under \textbf{Dominique Strauss-Kahn}, non-European countries were protesting, and critical voices were also raised from within the organisation.'' & 3 & \textcolor{darkgreen}{Accepted} & No specific events or non-generalizable characteristics are mentioned, and no additional entities are present. \\
\hline
``The EU’s reputation as a trusted guarantor of human rights and freedoms will be further compromised by \textbf{José Manuel Barroso}’s decision to meet one of the world’s most brutal dictators, Uzbek President Islam Karimov,'' remarks Galima Bukharbaeva, the editor of the independent Uzbek news site Uznews in an opinion piece published by Süddeutsche Zeitung.'' & 4 & \textcolor{darkgreen}{Accepted} & The relationship between José Manuel Barroso and Islam Karimov is contextualized within a ``meeting,'' which can be generalized to other entities. \\
\hline
``Even when Merkozy (\textbf{Angela Merkel} and Nicolas Sarkozy) were holding the reins, the balance was beginning to shift and the French were none too happy about it,'' explains Kęstutis Girnius.'' & 5 & \textcolor{burgundy}{Rejected} & ``Merkozy'' is a term specific to Angela Merkel and Nicolas Sarkozy and cannot be generalized. \\
\hline
``At the time, he also recommended that \textbf{Mussolini} and Hitler develop closer ties.'' & 6 & \textcolor{darkgreen}{Accepted} & The relationship between Hitler and Mussolini is not contextualized and can be generalized to other entities. \\
\hline
\end{tabular}
\end{table*}

This process ensures that only sentences free of counterfactual or overly specific information are used, allowing for meaningful and consistent entity substitution in downstream tasks.

\section{Sentence Adaptation and Translation}
\label{sentence_translation}

To ensure gender neutrality and adaptability across languages, we generate two versions of each sentence: one with male gender characteristics and one with female gender characteristics. This process was manually applied to all 450 sentences in our dataset. Additionally, to test political biases in non-Western languages tied to different geopolitical contexts, we translated the sentences into multiple languages. Below, we outline the steps taken to achieve this:

\paragraph{Step 1: Gender Adaptation}
\begin{itemize}
    \item For each sentence, we create two versions: one using male pronouns and characteristics (e.g., ``he,'' ``his'') and one using female pronouns and characteristics (e.g., ``she,'' ``her'').
\end{itemize}

\paragraph{Step 2: Preparing for Translation}
\begin{itemize}
    \item Before translating, we replace the placeholder ``X'' with a clearly gendered name (e.g., ``John'' for male, ``Mary'' for female) to ensure proper genderization in the target language.
\end{itemize}

\paragraph{Step 3: Translation}
\begin{itemize}
    \item Using DeepL, we translate the adapted sentences into the target languages (e.g., Arabic, French, Spanish, etc.).
\end{itemize}

\paragraph{Step 4: Recovering the Original Entity}
\begin{itemize}
    \item After translation, we replace the gendered name (e.g., ``Mary'') with the original placeholder to be replaced.
\end{itemize}

\paragraph{Step 5: Manual Verification}
\begin{itemize}
    \item To ensure accuracy, native speakers of each target language manually verify all translations. If some errors are found, they are manually corrected. This step confirms that no genderization errors or other inconsistencies were introduced during translation. 
\end{itemize}

\paragraph{Outcome}
\begin{itemize}
    \item We obtain 450 pairs of sentences (male and female versions) in six languages, ensuring gender adaptability and linguistic accuracy across diverse geopolitical contexts.
\end{itemize}

\begin{figure*}
    \centering
    \includegraphics[width=0.99\linewidth]{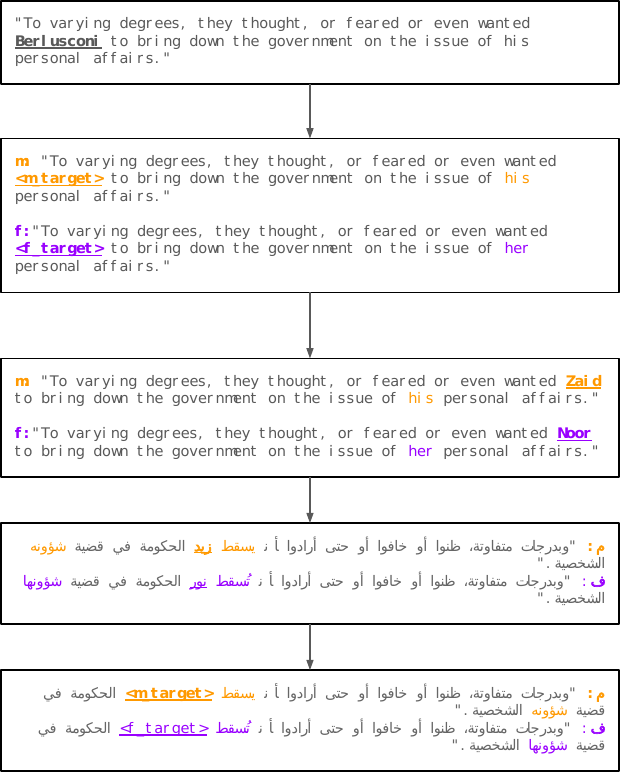}
    \caption{Steps for translating a sentence while ensuring gender adaptability. The process begins by creating two versions of the sentence with male and female gender characteristics. A clearly gendered name is then inserted in each version of the sentence to ensure proper genderization during translation. After translation, the introduced name is removed, and the placeholder is restored. Finally, all translated sentences are manually verified for accuracy and consistency of translation.}
    \label{fig:translation}
\end{figure*}

This process is summarized in Figure~\ref{fig:translation} with an example directly taken from the MAD-TSC sentences.

\section{Prompt Experiments}
\label{sec:prompt_experiments}

To identify the most effective prompt for sentiment analysis across all models, we experimented with three types of prompts: a 0-shot prompt, a 6-shot prompt, and a 9-shot prompt. The goal was to maximize performance across models while avoiding void answers or low accuracies that could compromise interpretability. We evaluated these prompts on the 450 selected sentences with the original entities in English (EN), French (FR), and Arabic (AR) using four models: Mistral 7B, Llama 3 8B, Qwen 2.5 7B, and Llama 3 70B. The results are summarized in Table~\ref{tab:prompt_results}.

\begin{table*}
\centering
\caption{F1-Scores for Prompt Experimentation Across Models and Languages}
\label{tab:prompt_results}
\begin{tabular}{cccccc}
\hline
\textbf{Prompt Type} & \textbf{Language} & \textbf{Mistral 7B} & \textbf{Llama 3 8B} & \textbf{Qwen 2.5 7B} & \textbf{Llama 3 70B} \\
\hline
0-shot & FR & 0.57 & 0.50 & 0.57 & 0.73 \\
       & EN & 0.65 & \textcolor{burgundy}{0.38} & 0.56 & 0.74 \\
       & AR & \textcolor{burgundy}{0.38} & \textcolor{burgundy}{0.43} & 0.64 & 0.64 \\
\hline
6-shot & FR & 0.61 & 0.64 & 0.69 & 0.72 \\
       & EN & 0.65 & 0.66 & 0.54 & 0.69 \\
       & AR & \textcolor{burgundy}{0.24} & \textcolor{burgundy}{0.41} & 0.63 & 0.69 \\
\hline
9-shot & FR & 0.66 & 0.66 & 0.67 & 0.74 \\
       & EN & 0.69 & 0.59 & 0.69 & 0.73 \\
       & AR & 0.53 & 0.63 & 0.62 & 0.71 \\
\hline
\end{tabular}
\end{table*}

\begin{figure*}
    \centering
    \includegraphics[width=0.99\linewidth]{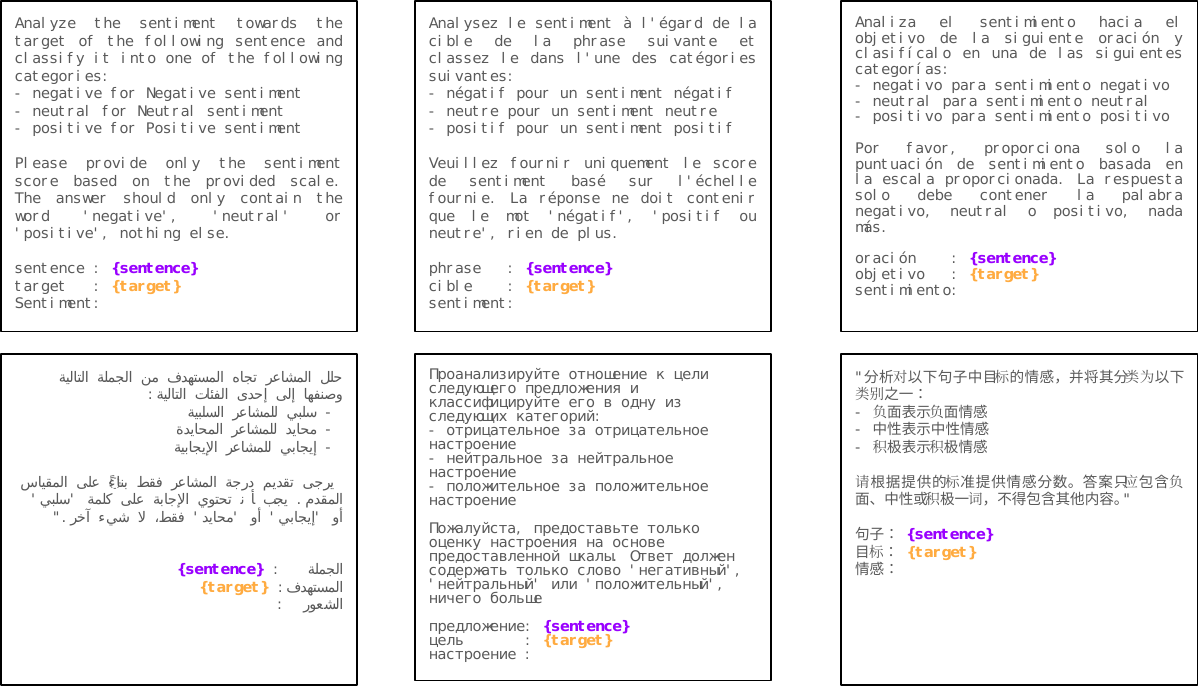}
    \caption{Prompts used for the experiment across six languages: English, French, and Spanish (first row), and Arabic, Russian, and Chinese (second row). For few-shot prompting, examples are added to the initial prompt along with their corresponding targets and sentiments. The order of examples is carefully randomized to ensure the LLM's output is not influenced by the sequence of sentiments provided.}
    \label{fig:prompts}
\end{figure*}

Based on these results, we selected the 9-shot prompt as it provided better performance across all languages, especially in Arabic, which had the lowest scores among the tested languages. The prompts are given in Figure \ref{fig:prompts}, and the few-shot examples are presented in Table \ref{tab:few_shot}.

Additionally, we tested variations where sentiment was expressed numerically ($-1$, $0$, $+1$) instead of verbally (\textit{negative}, \textit{neutral}, \textit{positive}), as well as constrained decoding (forcing models to select among predefined options). While these modifications did not improve performance, and in some cases worsened it, they yielded similar aggregated bias patterns. Given this, we opted not to include them in the final approach.

\begin{table*}[ht]
\centering
\caption{Sentences used for few-shot prompting.}
\label{tab:few_shot}
\renewcommand{\arraystretch}{1.2}
\begin{tabular}{|p{10cm}|c|c|}
\hline
\textbf{Sentence} & \textbf{Target} & \textbf{Sentiment} \\
\hline
“The outstanding experiences of my life,” he says, “are all bound up with the Vijećnica.” Very early in life, \textbf{Focak} came to love languages, literature, painting and architecture. & \textbf{Focak} & \textcolor{darkgreen}{Positive} \\
\hline
At sixteen, resembling a black-haired \textbf{Grace Kelly}, he devoured books on the architecture of the Renaissance and the works of Boccaccio and Dante in the library’s wood-paneled reading room. & \textbf{Grace Kelly} & \textcolor{darkgreen}{Positive} \\
\hline
\textbf{Merkel} has been the cork in the bottle with regard to tensions and populist powers in Europe. & \textbf{Merkel} & \textcolor{darkgreen}{Positive} \\
\hline
“He was incredibly brave” – muses \textbf{Bocheński} and adds that Kolakowski had set an example for democratic opposition in Poland. & \textbf{Bocheński} & Neutral \\
\hline
\textbf{Sterne} begins by pointing out that the IMF's analysis, which El-Erian correctly lauded, has been somewhat off target in Greece's case. & \textbf{Sterne} & Neutral \\
\hline
Facing a surprise rebellion from Mario Monti and \textbf{Mariano Rajoy}, she conceded crucial ground; she allowed the European Stability Mechanism (ESM) – that is the permanent European relief fund soon to be in place – to be able to capitalise Spanish banks directly and buy up Italian debt without requiring an austerity programme. & \textbf{Mariano Rajoy} & Neutral \\
\hline
In the run-up to the second round, the two contestants will attempt to win over protest voters and in particular the significant number that gave their backing to the discourse espoused by \textbf{Marine Le Pen}. & \textbf{Marine Le Pen} & \textcolor{burgundy}{Negative} \\
\hline
… all \textbf{Merkel} has to offer Monti’s Italy is words: words that are certainly new, but still only words. & \textbf{Merkel} & \textcolor{burgundy}{Negative} \\
\hline
Certainly, \textbf{Angela Merkel} speaks constantly of ‘European solidarity’, [...] but she is not ready to support young Greeks fleeing the crisis. & \textbf{Angela Merkel} & \textcolor{burgundy}{Negative} \\
\hline
\end{tabular}
\end{table*}

\section{Sentiment for Individual entities}

Some entities were particularly subject to much more positive or negative sentiment ratings. \textit{Giovanni Falcone} (\textcolor{darkgreen}{+0.14}), an Italian judge known for his work against organized crime, \textit{John F. Kennedy} (\textcolor{darkgreen}{+0.13}), the 35th U.S. president, \textit{Fannie Lou Hamer} (\textcolor{darkgreen}{+0.12}), an American civil rights activist, and \textit{Jacinda Ardern} (\textcolor{darkgreen}{+0.12}), former Prime Minister of New Zealand, were among the most highly rated entities. In contrast, \textit{Adolf Hitler} (\textcolor{burgundy}{-0.81}), \textit{Heinrich Himmler} (\textcolor{burgundy}{-0.55}), and \textit{Joseph Stalin} (\textcolor{burgundy}{-0.34}) received some of the most negative sentiment scores. \textit{Nick Griffin} (\textcolor{burgundy}{-0.36}), a British far-right politician and former leader of the British National Party, \textit{Marine Le Pen} (\textcolor{burgundy}{-0.28}), leader of France’s far-right National Rally party, and \textit{Donald Trump} (\textcolor{burgundy}{-0.23}) were also rated more negatively across models.

\section{Bias in Control Group}

Table \ref{tab:sentiments} shows the top fictional names linked to positive (\textcolor{darkgreen}{+}) and negative (\textcolor{burgundy}{-}) sentiments. Despite using fictional names to reduce bias, the results show clear patterns. Most names with positive sentiments are female, while those with negative sentiments are largely non-Western, with "Vladimir" being notably overrepresented. This indicates that bias persists. The overrepresentation of women in positive sentiments may contribute to a slight positive bias in some cases, while the prevalence of non-Western names in negative sentiments introduces a slight negative bias in others. For instance, Russian names are overrepresented in the FL alignment while women are overrepresented in the LL alignment. These findings highlight the difficulty of achieving complete neutrality through fictional name generation. 

\begin{table*}
\centering
\caption{Top Positive and Negative Sentiments associated with fictional names.}
\label{tab:sentiments}
\begin{tabular}{l|c}
\hline
\textbf{Fictional Name} & \textbf{Sentiment} \\ \hline
Talita Ribeiro & \textcolor{darkgreen}{+0.0364} \\
Miyu Hoshikawa & \textcolor{darkgreen}{+0.0357} \\
Alyssa Verhaegen & \textcolor{darkgreen}{+0.0350} \\
Anushri Desai & \textcolor{darkgreen}{+0.0335} \\
Dante Mirabello & \textcolor{darkgreen}{+0.0335} \\
Celina Salcedo & \textcolor{darkgreen}{+0.0326} \\
Shanvika Mehta & \textcolor{darkgreen}{+0.0323} \\
Eliana Cardozo & \textcolor{darkgreen}{+0.0316} \\
Anaïs Rochelet & \textcolor{darkgreen}{+0.0313} \\
Milena Cortés & \textcolor{darkgreen}{+0.0312} \\ \hline
Boris Gromov & \textcolor{burgundy}{-0.0363} \\
Vladimir Vasiliev & \textcolor{burgundy}{-0.0368} \\
Vladimir Khamatov & \textcolor{burgundy}{-0.0370} \\
Lázaro Trujillo & \textcolor{burgundy}{-0.0409} \\
Stepan Lysenko & \textcolor{burgundy}{-0.0429} \\
Reginald Tyndall & \textcolor{burgundy}{-0.0439} \\
Boris Aksyonov & \textcolor{burgundy}{-0.0492} \\
Miklóska Szalaj & \textcolor{burgundy}{-0.0545} \\
Vladimir Antonov & \textcolor{burgundy}{-0.0571} \\
János Sárközi & \textcolor{burgundy}{-0.0571} \\ \hline
\end{tabular}
\end{table*}

\section{Statistical Tests}
\label{sec_stat_tests}
To assess potential biases in sentiment alignment, we conduct Mann-Whitney U tests between alignment values for each model-language pair. Let $\mu_x$ denote the mean sentiment of an observation $X$ belonging to alignment $A_1$, and $\mu_y$ denote the mean sentiment of an observation $Y$ belonging to alignment $A_2$. The null hypothesis $H_0$ is formulated as:
\[
H_0: P(\mu_y > \mu_x) = 0.5,
\]
implying no systematic difference in sentiment between alignments $A_1$ and $A_2$. Rejection of $H_0$ suggests that $P(\mu_y > \mu_x) \neq 0.5$, indicating a significant bias towards one alignment over the other. Specifically, if $H_0$ is rejected and $P(\mu_y > \mu_x) > 0.5$, this suggests a bias towards alignment $A_2$ compared to $A_1$.

We report the results of these tests in tables for each model-language pair. Each cell in the table contains the $p$-value for the Mann-Whitney U test comparing the sentiment distributions of alignment $i$ (row) and alignment $j$ (column). A $p$-value $< 0.01$ indicates a statistically significant bias towards the alignment in the row compared to the alignment in the column. For example, in Table \ref{tab:aya} for French, the value in row \textbf{LL} and column \textbf{FL} is $0.0029$ ($< 0.01$). This means that, for the model Aya-Expanse-32B, there is a significant bias favoring \textbf{LL} over \textbf{FL} in French. This approach allows us to systematically identify and quantify biases across different alignments and languages.

Overall, the results show a clear bias in favor of left, center-left, and center alignments, while revealing a consistent negative bias towards right and far-right alignments. This pattern highlights systematic disparities in sentiment alignment across the political spectrum.


\begin{table*}
\centering
\caption{Model cards for the evaluated LLMs.}
\label{tab:model_cards}
\renewcommand{\arraystretch}{1.5} 
\resizebox{\textwidth}{!}{ 
\begin{tabular}{lccccc}
\hline
\textbf{Name} & \textbf{Full Name} & \textbf{Company} & \textbf{Country} & \textbf{Size} & \textbf{Release Date} \\ \hline
\multirow{2}{*}{\textbf{Qwen-7B}} & \multirow{2}{*}{Qwen-7B-Instruct} & \multirow{2}{*}{Alibaba} & \multirow{2}{*}{China} & 7B & \multirow{2}{*}{2023} \\ 
\textbf{Qwen-72B} & Qwen-72B-Instruct & & & 72B & \\ \hline
\multirow{2}{*}{\textbf{Llama 3-8B}} & \multirow{2}{*}{Meta-Llama-3-8B-Instruct} & \multirow{2}{*}{Meta} & \multirow{2}{*}{USA} & 8B & \multirow{2}{*}{2024} \\ 
\textbf{Llama 3-70B} & Meta-Llama-3-70B-Instruct & & & 70B & \\ \hline
\textbf{Mistral-7B} & Mistral-7B-Instruct-v.03 & Mistral AI & France & 7B & 2023 \\ \hline
\textbf{Aya-Expanse-32B} & Aya-Expanse-32B & Cohere & Canada & 32B & 2023 \\ \hline
\textbf{GPT-4o-mini} & GPT-4o-mini & OpenAI & USA & Unknown & 2023 \\ \hline
\end{tabular}
}
\end{table*}

\begin{table*}
\centering
\caption{Aya-Expanse-32B: Mann-Withney test p-values.}
\label{tab:aya}
\begin{tabular}{l|l|ccccccccc}
\hline
\textbf{Model} & \textbf{Language} &  & \textbf{FL} & \textbf{LL} & \textbf{CL} & \textbf{CC} & \textbf{CR} & \textbf{RR} & \textbf{FR} & \textbf{BT} \\ \hline
\multirow{9}{*}{\textbf{Aya}} & \multirow{9}{*}{French} & \textbf{FL} & & 0.9972 & 0.9960 & 0.8845 & 0.0012 & 0.0000 & 0.0000 & 0.2945 \\
 & & \textbf{LL} & 0.0029 &  & 0.4070 & 0.0641 & 0.0000 & 0.0000 & 0.0000 & 0.0017\\
 & & \textbf{CL} & 0.0040 & 0.5934 &  & 0.0878 & 0.0000 & 0.0000 & 0.0000 & 0.0028\\
 & & \textbf{CC} & 0.1157 & 0.9361 & 0.9123 &  & 0.0000 & 0.0000 & 0.0000 & 0.0520 \\
 & & \textbf{CR} & 0.9988 & 1.0000 & 1.0000 & 1.0000 &  & 0.0021 & 0.0000 & 0.9870\\
 & & \textbf{RR} & 1.0000 & 1.0000 & 1.0000 & 1.0000 & 0.9979 &  & 0.0006 & 1.0000\\
 & & \textbf{FR} & 1.0000 & 1.0000 & 1.0000 & 1.0000 & 1.0000 & 0.9994 &   & 1.0000 \\
 & & \textbf{BT} & 0.7060 & 0.9983 & 0.9972 & 0.9482 & 0.0131 & 0.0000 & 0.0000 \\ \hline
\multirow{9}{*}{\textbf{Aya}} & \multirow{9}{*}{English} & \textbf{FL} & & 0.9857 & 0.8960 & 0.8011 & 0.0033 & 0.0000 & 0.0000 & 0.2036 \\
 & & \textbf{LL} & 0.0144 &  & 0.1468 & 0.0798 & 0.0000 & 0.0000 & 0.0000 & 0.0031 \\
 & & \textbf{CL} & 0.1042 & 0.8534 &  & 0.3406 & 0.0000 & 0.0000 & 0.0000 & 0.0214 \\
 & & \textbf{CC} & 0.1992 & 0.9203 & 0.6598 &  & 0.0001 & 0.0000 & 0.0000 & 0.0537 \\
 & & \textbf{CR} & 0.9967 & 1.0000 & 1.0000 & 0.9999 &  & 0.0006 & 0.0000 & 0.9355 \\
 & & \textbf{RR} & 1.0000 & 1.0000 & 1.0000 & 1.0000 & 0.9994 &  & 0.0036 & 1.0000 \\
 & & \textbf{FR} & 1.0000 & 1.0000 & 1.0000 & 1.0000 & 1.0000 & 0.9964 &  & 1.0000 \\
 & & \textbf{BT} & 0.7969 & 0.9970 & 0.9786 & 0.9465 & 0.0647 & 0.0000 & 0.0000 &  \\ \hline

 \multirow{9}{*}{\textbf{Aya}} & \multirow{9}{*}{Spanish} & \textbf{FL} & & 0.9749& 	0.8956& 	0.9329&	0.0062&	0.0000&	0.0000	&0.0947
\\
 & & \textbf{LL} & 0.0252&	&	0.2223&	0.3258&	0.0000&	0.0000	&0.0000&	0.0010 \\
 & & \textbf{CL} & 0.1046&	0.7780	&	&0.6325	&0.0000	&0.0000&	0.0000	&0.0048\\
 & & \textbf{CC}  & 0.0673&	0.6746	&0.3679	&	&0.0000	&0.0000	&0.0000	&0.0025  \\
 & & \textbf{CR} &  0.9938&	1.0000	&1.0000&	1.0000	&	&0.0011	&0.0000	&0.8047 \\
 & & \textbf{RR} &1.0000	&1.0000	&1.0000&	1.0000	&0.9989	&	&0.0058	&0.9994 \\
 & & \textbf{FR} &1.0000	&1.0000	&1.0000	&1.0000	&1.0000	&0.9942	&	&1.0000 \\
 & & \textbf{BT} &0.9056&	0.9991	&0.9952	&0.9975	&0.1958&	0.0006	&0.0000&  \\ \hline

  \multirow{9}{*}{\textbf{Aya}} & \multirow{9}{*}{Russian} & \textbf{FL} & &0.9837	&0.9802	&0.7870	&0.3660&	0.0005	&0.0000	&0.5584
\\
 & & \textbf{LL} & 0.0163&		&0.4237	&0.0952	&0.0085	&0.0000	&0.0000	&0.0436\\
 & & \textbf{CL} & 0.0199	&0.5766	&	&0.1206	&0.0086	&0.0000	&0.0000	&0.0478 \\
 & & \textbf{CC}  &0.2133	&0.9050	&0.8796	&	&0.1329	&0.0000	&0.0000	&0.2568 \\
 & & \textbf{CR} & 0.6344	&0.9916	&0.9914	&0.8674&		&0.0011	&0.0000	&0.6948\\
 & & \textbf{RR} &0.9995	&1.0000	&1.0000	&1.0000	&0.9989	&	&0.0518	&0.9992\\
 & & \textbf{FR} &1.0000	&1.0000&	1.0000	&1.0000	&1.0000	&0.9483	&	&1.0000\\
 & & \textbf{BT} &0.4423	&0.9566	&0.9523&	0.7436	&0.3058	&0.0008&	0.0000
&  \\ \hline

  \multirow{9}{*}{\textbf{Aya}} & \multirow{9}{*}{Chinese} & \textbf{FL} & &0.9927	&0.9884	&0.9820	&0.6715	&0.0572	&0.0000	&0.5017
\\
 & & \textbf{LL} & 0.0074&		&0.4157	&0.4082&	0.0267	&0.0000	&0.0000	&0.0164\\
 & & \textbf{CL} &0.0116	&0.5847	&	&0.4754	&0.0378	&0.0001	&0.0000	&0.0256 \\
 & & \textbf{CC} & 0.0181&	0.5922	&0.5250	&	&0.0544	&0.0001	&0.0000&	0.0266 \\
 & & \textbf{CR}  &0.3289	&0.9734&	0.9623	&0.9457	&	&0.0198	&0.0000	&0.3019\\
 & & \textbf{RR} &0.9429	&1.0000	&0.9999	&0.9999	&0.9803	&	&0.0077	&0.9128\\
 & & \textbf{FR} &1.0000	&1.0000	&1.0000	&1.0000	&1.0000	&0.9923	&	&0.9996\\
 & & \textbf{BT} &0.4990&	0.9837	&0.9744	&0.9735	&0.6986	&0.0875	&0.0004
&  \\ \hline

\multirow{9}{*}{\textbf{Aya}} & \multirow{9}{*}{Arabic} & \textbf{FL} & &0.9737	&0.9965	&0.9109	&0.5798	&0.0074	&0.0000	&0.3467
\\
 & & \textbf{LL} & 0.0264&		&0.7777	&0.1758	&0.0256	&0.0000	&0.0000	&0.0080\\
 & & \textbf{CL} & 0.0035	&0.2226	&	&0.0564	&0.0040	&0.0000	&0.0000	&0.0013 \\
 & & \textbf{CC}  & 0.0893&	0.8245	&0.9437	&	&0.1244	&0.0000	&0.0000&	0.0531 \\
 & & \textbf{CR} & 0.4207	&0.9745	&0.9960	&0.8758	&	&0.0028	&0.0000	&0.2835\\
 & & \textbf{RR} & 0.9926	&1.0000	&1.0000	&1.0000	&0.9972	&	&0.0023	&0.9717\\
 & & \textbf{FR} & 1.0000	&1.0000	&1.0000	&1.0000	&1.0000	&0.9977	&	&1.0000\\
 & & \textbf{BT} & 0.6539	&0.9921	&0.9987	&0.9470	&0.7170	&0.0284	&0.0000
&  \\ \hline

\end{tabular}
\end{table*}

\begin{table*}
\centering
\caption{GPT-4o-mini: Mann-Withney test p-values.}
\label{tab:chatgpt}
\begin{tabular}{l|l|ccccccccc}
\hline
\textbf{Model} & \textbf{Language} &  & \textbf{FL} & \textbf{LL} & \textbf{CL} & \textbf{CC} & \textbf{CR} & \textbf{RR} & \textbf{FR} & \textbf{BT} \\ \hline

\multirow{9}{*}{\textbf{GPT}} & \multirow{9}{*}{French} & \textbf{FL} & &0.8329	&0.0162	&0.0515	&0.0000	&0.0000	&0.0000	&0.0067
\\
 & & \textbf{LL} & 0.1673&		&0.0005	&0.0034	&0.0000	&0.0000	&0.0000	&0.0004\\
 & & \textbf{CL} & 0.9839	&0.9995	&	&0.5517	&0.0001	&0.0000	&0.0000	&0.1251 \\
 & & \textbf{CC}  & 0.9486	&0.9966	&0.4487	&	&0.0002	&0.0000	&0.0000&	0.1231 \\
 & & \textbf{CR} & 1.0000	&1.0000	&0.9999	&0.9998	&	&0.0002	&0.0000	&0.9688\\
 & & \textbf{RR} & 1.0000	&1.0000	&1.0000	&1.0000	&0.9998	&	&0.1928	&1.0000\\
 & & \textbf{FR} & 1.0000	&1.0000	&1.0000	&1.0000	&1.0000	&0.8075	&	&1.0000\\
 & & \textbf{BT} & 0.9933	&0.9996	&0.8752	&0.8773	&0.0313	&0.0000	&0.0000
&  \\ \hline

\multirow{9}{*}{\textbf{GPT}} & \multirow{9}{*}{English} & \textbf{FL} & &0.8267	&0.2950	&0.0311	&0.0000	&0.0000	&0.0015	&0.0571
\\
 & & \textbf{LL} & 0.1736&		&0.0421	&0.0022	&0.0000	&0.0000	&0.0000	&0.0050\\
 & & \textbf{CL} & 0.7054	&0.9580	&	&0.0893	&0.0001	&0.0000	&0.0066	&0.1180 \\
 & & \textbf{CC}  & 0.9689	&0.9979	&0.9108	&	&0.0073	&0.0000	&0.1221&	0.4831 \\
 & & \textbf{CR} & 1.0000	&1.0000	&0.9999	&0.9928	&	&0.0271	&0.9146	&0.9815\\
 & & \textbf{RR} & 1.0000	&1.0000	&1.0000	&1.0000	&0.9730	&	&0.9994	&0.9999\\
 & & \textbf{FR} & 0.9985	&1.0000	&0.9934	&0.8781	&0.0856	&0.0006	&	&0.8210\\
 & & \textbf{BT} & 0.9431	&0.9950	&0.8822	&0.5176	&0.0185	&0.0001	&0.1794
&  \\ \hline

\multirow{9}{*}{\textbf{GPT}} & \multirow{9}{*}{Spanish} & \textbf{FL} & &0.9727	&0.0656	&0.0565	&0.0000	&0.0000	&0.0000	&0.0035
\\
 & & \textbf{LL} & 0.0273&		&0.0001	&0.0002	&0.0000	&0.0000	&0.0000	&0.0000\\
 & & \textbf{CL} & 0.9345	&0.9999	&	&0.4126	&0.0004	&0.0000	&0.0000	&0.0445 \\
 & & \textbf{CC}  & 0.9436	&0.9998	&0.5878	&	&0.0020	&0.0000	&0.0000&	0.0683 \\
 & & \textbf{CR} & 1.0000	&1.0000	&0.9996	&0.9980	&	&0.0003	&0.0000	&0.8600\\
 & & \textbf{RR} & 1.0000	&1.0000	&1.0000	&1.0000	&0.9997	&	&0.0908	&1.0000\\
 & & \textbf{FR} & 1.0000	&1.0000	&1.0000	&1.0000	&1.0000	&0.9093	&	&1.0000\\
 & & \textbf{BT} & 0.9965	&1.0000	&0.9556	&0.9319	&0.1404	&0.0000	&0.0000
&  \\ \hline

\multirow{9}{*}{\textbf{GPT}} & \multirow{9}{*}{Russian} & \textbf{FL} & &0.9220	&0.3050	&0.2979	&0.0014	&0.0002	&0.0040	&0.0247
\\
 & & \textbf{LL} & 0.0782&		&0.0183	&0.0238	&0.0000	&0.0000	&0.0000	&0.0004\\
 & & \textbf{CL} & 0.6954	&0.9818	&	&0.4998	&0.0047	&0.0006	&0.0095	&0.0525 \\
 & & \textbf{CC}  & 0.7025	&0.9762	&0.5006	&	&0.0058	&0.0009	&0.0153&	0.0679 \\
 & & \textbf{CR} & 0.9986	&1.0000	&0.9954	&0.9942	&	&0.2527	&0.6423	&0.7616\\
 & & \textbf{RR} & 0.9998	&1.0000	&0.9994	&0.9991	&0.7476	&	&0.8554	&0.8964\\
 & & \textbf{FR} & 0.9960	&1.0000	&0.9905	&0.9847	&0.3581	&0.1449	&	&0.6287\\
 & & \textbf{BT} & 0.9754	&0.9996	&0.9477	&0.9323	&0.2389	&0.1039	&0.3719
&  \\ \hline

\multirow{9}{*}{\textbf{GPT}} & \multirow{9}{*}{Chinese} & \textbf{FL} & &0.8267	&0.2950	&0.0311	&0.0000	&0.0000	&0.0015	&0.0571
\\
 & & \textbf{LL} & 0.1736&		&0.0421	&0.0022	&0.0000	&0.0000	&0.0000	&0.0050\\
 & & \textbf{CL} & 0.7054	&0.9580	&	&0.0893	&0.0001	&0.0000	&0.0066	&0.1180 \\
 & & \textbf{CC}  & 0.9689	&0.9979	&0.9108	&	&0.0073	&0.0000	&0.1221&	0.4831 \\
 & & \textbf{CR} & 1.0000	&1.0000	&0.9999	&0.9928	&	&0.0271	&0.9146	&0.9815\\
 & & \textbf{RR} & 1.0000	&1.0000	&1.0000	&1.0000	&0.9730	&	&0.9994	&0.9999\\
 & & \textbf{FR} & 0.9985	&1.0000	&0.9934	&0.8781	&0.0856	&0.0006	&	&0.8210\\
 & & \textbf{BT} & 0.9431	&0.9950	&0.8822	&0.5176	&0.0185	&0.0001	&0.1794
&  \\ \hline

\multirow{9}{*}{\textbf{GPT}} & \multirow{9}{*}{Arabic} & \textbf{FL} & &0.9187	&0.3627	&0.3310	&0.0232	&0.0000	&0.0022	&0.0257
\\
 & & \textbf{LL} & 0.0814&		&0.0264	&0.0257	&0.0002	&0.0000	&0.0000	&0.0006\\
 & & \textbf{CL} & 0.6377	&0.9737	&	&0.4423	&0.0309	&0.0000	&0.0037	&0.0317 \\
 & & \textbf{CC}  & 0.6694	&0.9744	&0.5580	&	&0.0492	&0.0000	&0.0062&	0.0503 \\
 & & \textbf{CR} & 0.9768	&0.9998	&0.9691	&0.9509	&	&0.0007	&0.2089	&0.4375\\
 & & \textbf{RR} & 1.0000	&1.0000	&1.0000	&1.0000	&0.9993	&	&0.9926	&0.9967\\
 & & \textbf{FR} & 0.9978	&1.0000	&0.9963	&0.9938	&0.7914	&0.0074	&	&0.7009\\
 & & \textbf{BT} & 0.9744	&0.9994	&0.9684	&0.9498	&0.5632	&0.0033	&0.2997
&  \\ \hline

\end{tabular}
\end{table*}

\begin{table*}
\centering
\caption{Llama 3-70B: Mann-Withney test p-values.}
\label{tab:llama70b}
\begin{tabular}{l|l|ccccccccc}
\hline
\textbf{Model} & \textbf{Language} &  & \textbf{FL} & \textbf{LL} & \textbf{CL} & \textbf{CC} & \textbf{CR} & \textbf{RR} & \textbf{FR} & \textbf{BT} \\ \hline

\multirow{9}{*}{\textbf{Llama-70b}} & \multirow{9}{*}{French} & \textbf{FL} & &1.0000	&1.0000	&0.9977	&0.4640	&0.0000	&0.0000	&0.3502
\\
 & & \textbf{LL} & 0.0000&		&0.5633	&0.0540	&0.0000	&0.0000	&0.0000	&0.0000\\
 & & \textbf{CL} & 0.0000	&0.4371	&	&0.0270	&0.0000	&0.0000	&0.0000	&0.0000 \\
 & & \textbf{CC}  & 0.0023	&0.9461	&0.9730	&	&0.0014	&0.0000	&0.0000&	0.0032 \\
 & & \textbf{CR} & 0.5365	&1.0000	&1.0000	&0.9986	&	&0.0000	&0.0000	&0.3803\\
 & & \textbf{RR} & 1.0000	&1.0000	&1.0000	&1.0000	&1.0000	&	&0.0008	&1.0000\\
 & & \textbf{FR} & 1.0000	&1.0000	&1.0000	&1.0000	&1.0000	&0.9992	&	&1.0000\\
 & & \textbf{BT} & 0.6504	&1.0000	&1.0000	&0.9968	&0.6203	&0.0000	&0.0000
&  \\ \hline

\multirow{9}{*}{\textbf{Llama-70b}} & \multirow{9}{*}{English} & \textbf{FL} & &1.0000	&1.0000	&0.9991	&0.2828	&0.0000	&0.0000	&0.1904
\\
 & & \textbf{LL} & 0.0000&		&0.5504	&0.0908	&0.0000	&0.0000	&0.0000	&0.0000\\
 & & \textbf{CL} & 0.0000	&0.4499	&	&0.0569	&0.0000	&0.0000	&0.0000	&0.0000 \\
 & & \textbf{CC}  & 0.0009	&0.9093	&0.9433	&	&0.0001	&0.0000	&0.0000&	0.0002 \\
 & & \textbf{CR} & 0.7176	&1.0000	&1.0000	&0.9999	&	&0.0000	&0.0000	&0.2787\\
 & & \textbf{RR} & 1.0000	&1.0000	&1.0000	&1.0000	&1.0000	&	&0.0013	&1.0000\\
 & & \textbf{FR} & 1.0000	&1.0000	&1.0000	&1.0000	&1.0000	&0.9987	&	&1.0000\\
 & & \textbf{BT} & 0.8101	&1.0000	&1.0000	&0.9998	&0.7218	&0.0000	&0.0000
&  \\ \hline

\multirow{9}{*}{\textbf{Llama-70b}} & \multirow{9}{*}{Spanish} & \textbf{FL} & &1.0000	&1.0000	&0.9994	&0.7008	&0.0000	&0.0000	&0.2141
\\
 & & \textbf{LL} & 0.0000&		&0.5510	&0.1371	&0.0000	&0.0000	&0.0000	&0.0000\\
 & & \textbf{CL} & 0.0000	&0.4494	&	&0.0962	&0.0000	&0.0000	&0.0000	&0.0000 \\
 & & \textbf{CC}  & 0.0006	&0.8631	&0.9040	&	&0.0019	&0.0000	&0.0000&	0.0001 \\
 & & \textbf{CR} & 0.2996	&1.0000	&1.0000	&0.9981	&	&0.0000	&0.0000	&0.1049\\
 & & \textbf{RR} & 1.0000	&1.0000	&1.0000	&1.0000	&1.0000	&	&0.0000	&0.9997\\
 & & \textbf{FR} & 1.0000	&1.0000	&1.0000	&1.0000	&1.0000	&1.0000	&	&1.0000\\
 & & \textbf{BT} & 0.7864	&1.0000	&1.0000	&0.9999	&0.8954	&0.0003	&0.0000
&  \\ \hline

\multirow{9}{*}{\textbf{Llama-70b}} & \multirow{9}{*}{Russian} & \textbf{FL} & &0.9730	&0.9830	&0.9269	&0.6134	&0.0075	&0.0001	&0.2847
\\
 & & \textbf{LL} & 0.0270&		&0.5975	&0.3040	&0.0433	&0.0000	&0.0000	&0.0094\\
 & & \textbf{CL} & 0.0171	&0.4029	&	&0.2201	&0.0298	&0.0000	&0.0000	&0.0059 \\
 & & \textbf{CC}  & 0.0732	&0.6963	&0.7801	&	&0.1019	&0.0000	&0.0000&	0.0235 \\
 & & \textbf{CR} & 0.3871	&0.9568	&0.9702	&0.8983	&	&0.0018	&0.0000	&0.1804\\
 & & \textbf{RR} & 0.9925	&1.0000	&1.0000	&1.0000	&0.9982	&	&0.0719	&0.9518\\
 & & \textbf{FR} & 0.9999	&1.0000	&1.0000	&1.0000	&1.0000	&0.9283	&	&0.9974\\
 & & \textbf{BT} & 0.7159	&0.9906	&0.9941	&0.9766	&0.8200	&0.0484	&0.0026
&  \\ \hline

\multirow{9}{*}{\textbf{Llama-70b}} & \multirow{9}{*}{Chinese} & \textbf{FL} & &0.8008	&0.7796	&0.5923	&0.3158	&0.0150	&0.0001	&0.7929
\\
 & & \textbf{LL} & 0.1995&		&0.4301	&0.2680	&0.0831	&0.0007	&0.0000	&0.4873\\
 & & \textbf{CL} & 0.2207	&0.5703	&	&0.3468	&0.1253	&0.0019	&0.0000	&0.5374 \\
 & & \textbf{CC}  & 0.4081	&0.7323	&0.6535	&	&0.2120	&0.0044	&0.0000&	0.7106 \\
 & & \textbf{CR} & 0.6847	&0.9171	&0.8750	&0.7883	&	&0.0409	&0.0005	&0.8663\\
 & & \textbf{RR} & 0.9850	&0.9993	&0.9981	&0.9956	&0.9592	&	&0.0606	&0.9939\\
 & & \textbf{FR} & 0.9999	&1.0000	&1.0000	&1.0000	&0.9995	&0.9396	&	&1.0000\\
 & & \textbf{BT} & 0.2076	&0.5133	&0.4632	&0.2899	&0.1341	&0.0061	&0.0000
&  \\ \hline

\multirow{9}{*}{\textbf{Llama-70b}} & \multirow{9}{*}{Arabic} & \textbf{FL} & &0.9980	&0.9998	&0.9868	&0.6618	&0.2713	&0.0031	&0.7860
\\
 & & \textbf{LL} & 0.0020&		&0.7030	&0.2030	&0.0032	&0.0000	&0.0000	&0.0348\\
 & & \textbf{CL} & 0.0002	&0.2973	&	&0.0849	&0.0006	&0.0000	&0.0000	&0.0145 \\
 & & \textbf{CC}  & 0.0133	&0.7972	&0.9152	&	&0.0367	&0.0016	&0.0000&	0.1574 \\
 & & \textbf{CR} & 0.3386	&0.9968	&0.9994	&0.9634	&	&0.1303	&0.0003	&0.7260\\
 & & \textbf{RR} & 0.7291	&1.0000	&1.0000	&0.9985	&0.8700	&	&0.0086	&0.9410\\
 & & \textbf{FR} & 0.9969	&1.0000	&1.0000	&1.0000	&0.9997	&0.9914	&	&0.9998\\
 & & \textbf{BT} & 0.2145	&0.9653	&0.9855	&0.8430	&0.2745	&0.0592	&0.0002
&  \\ \hline

\end{tabular}
\end{table*}

\begin{table*}
\centering
\caption{Llama 3-8B: Mann-Withney test p-values.}
\label{tab:llama-8b}
\begin{tabular}{l|l|ccccccccc}
\hline
\textbf{Model} & \textbf{Language} &  & \textbf{FL} & \textbf{LL} & \textbf{CL} & \textbf{CC} & \textbf{CR} & \textbf{RR} & \textbf{FR} & \textbf{BT} \\ \hline

\multirow{9}{*}{\textbf{Llama-8B}} & \multirow{9}{*}{French} & \textbf{FL} & &0.8490	&0.6812	&0.7593	&0.0068	&0.0000	&0.0000	&0.1432
\\
 & & \textbf{LL} & 0.1512&		&0.2606	&0.3308	&0.0001	&0.0000	&0.0000	&0.0229\\
 & & \textbf{CL} & 0.3191	&0.7397	&	&0.6266	&0.0009	&0.0000	&0.0000	&0.0664 \\
 & & \textbf{CC}  & 0.2410	&0.6696	&0.3738	&	&0.0003	&0.0000	&0.0000&	0.0412 \\
 & & \textbf{CR} & 0.9932	&0.9999	&0.9991	&0.9997	&	&0.0001	&0.0010	&0.8457\\
 & & \textbf{RR} & 1.0000	&1.0000	&1.0000	&1.0000	&0.9999	&	&0.5110	&1.0000\\
 & & \textbf{FR} & 1.0000	&1.0000	&1.0000	&1.0000	&0.9990	&0.4894	&	&0.9999\\
 & & \textbf{BT} & 0.8571	&0.9772	&0.9338	&0.9589	&0.1547	&0.0000	&0.0001
&  \\ \hline

\multirow{9}{*}{\textbf{Llama-8B}} & \multirow{9}{*}{English} & \textbf{FL} & &0.9947	&0.9502	&0.9546	&0.3270	&0.0005	&0.0148	&0.6682
\\
 & & \textbf{LL} & 0.0053&		&0.1141	&0.1381	&0.0005	&0.0000	&0.0000	&0.0402\\
 & & \textbf{CL} & 0.0499	&0.8861	&	&0.6259	&0.0140	&0.0000	&0.0001	&0.1664 \\
 & & \textbf{CC}  & 0.0455	&0.8622	&0.3745	&	&0.0069	&0.0000	&0.0000&	0.1418 \\
 & & \textbf{CR} & 0.6734	&0.9995	&0.9860	&0.9931	&	&0.0009	&0.0316	&0.7830\\
 & & \textbf{RR} & 0.9995	&1.0000	&1.0000	&1.0000	&0.9991	&	&0.7867	&0.9993\\
 & & \textbf{FR} & 0.9852	&1.0000	&0.9999	&1.0000	&0.9685	&0.2136	&	&0.9888\\
 & & \textbf{BT} & 0.3325	&0.9599	&0.8339	&0.8585	&0.2175	&0.0007	&0.0113
&  \\ \hline

\multirow{9}{*}{\textbf{Llama-8B}} & \multirow{9}{*}{Spanish} & \textbf{FL} & &0.9955	&0.9196	&0.7252	&0.0065	&0.0000	&0.0000	&0.5098
\\
 & & \textbf{LL} & 0.0045&		&0.0965	&0.0129	&0.0000	&0.0000	&0.0000	&0.0225\\
 & & \textbf{CL} & 0.0806	&0.9036	&	&0.2008	&0.0000	&0.0000	&0.0000	&0.1480 \\
 & & \textbf{CC}  & 0.2752	&0.9871	&0.7995	&	&0.0005	&0.0000	&0.0000&	0.3308 \\
 & & \textbf{CR} & 0.9935	&1.0000	&1.0000	&0.9995	&	&0.0001	&0.0016	&0.9866\\
 & & \textbf{RR} & 1.0000	&1.0000	&1.0000	&1.0000	&0.9999	&	&0.6245	&1.0000\\
 & & \textbf{FR} & 1.0000	&1.0000	&1.0000	&1.0000	&0.9984	&0.3759	&	&1.0000\\
 & & \textbf{BT} & 0.4909	&0.9776	&0.8523	&0.6697	&0.0135	&0.0000	&0.0000
&  \\ \hline

\multirow{9}{*}{\textbf{Llama-8B}} & \multirow{9}{*}{Russian} & \textbf{FL} & &0.9995	&0.9542	&0.9961	&0.7795	&0.5049	&0.3953	&0.9251
\\
 & & \textbf{LL} & 0.0005&		&0.0586	&0.2616	&0.0056	&0.0009	&0.0002	&0.1142\\
 & & \textbf{CL} & 0.0459	&0.9415	&	&0.8272	&0.1814	&0.0498	&0.0260	&0.5048 \\
 & & \textbf{CC}  & 0.0039	&0.7387	&0.1730	&	&0.0390	&0.0052	&0.0023&	0.2490 \\
 & & \textbf{CR} & 0.2209	&0.9944	&0.8188	&0.9611	&	&0.2618	&0.1639	&0.7748\\
 & & \textbf{RR} & 0.4956	&0.9991	&0.9503	&0.9949	&0.7385	&	&0.3572	&0.9161\\
 & & \textbf{FR} & 0.6051	&0.9998	&0.9741	&0.9977	&0.8364	&0.6432	&	&0.9440\\
 & & \textbf{BT} & 0.0751	&0.8861	&0.4958	&0.7515	&0.2257	&0.0842	&0.0561
&  \\ \hline

\multirow{9}{*}{\textbf{Llama-8B}} & \multirow{9}{*}{Chinese} & \textbf{FL} & &0.9824	&0.9923	&0.9582	&0.3158	&0.0673	&0.1180	&0.7638
\\
 & & \textbf{LL} & 0.0176&		&0.6185	&0.3327	&0.0041	&0.0002	&0.0005	&0.1521\\
 & & \textbf{CL} & 0.0077	&0.3819	&	&0.2489	&0.0012	&0.0000	&0.0001	&0.0970 \\
 & & \textbf{CC}  & 0.0419	&0.6677	&0.7514	&	&0.0107	&0.0007	&0.0016&	0.2326 \\
 & & \textbf{CR} & 0.6846	&0.9959	&0.9988	&0.9893	&	&0.1640	&0.2367	&0.8821\\
 & & \textbf{RR} & 0.9328	&0.9998	&1.0000	&0.9993	&0.8363	&	&0.5772	&0.9763\\
 & & \textbf{FR} & 0.8822	&0.9995	&0.9999	&0.9984	&0.7637	&0.4232	&	&0.9623\\
 & & \textbf{BT} & 0.2367	&0.8483	&0.9032	&0.7679	&0.1182	&0.0238	&0.0379
&  \\ \hline

\end{tabular}
\end{table*}

\begin{table*}
\centering
\caption{Mistral-7B: Mann-Withney test p-values.}
\label{tab:mistral}
\begin{tabular}{l|l|ccccccccc}
\hline
\textbf{Model} & \textbf{Language} &  & \textbf{FL} & \textbf{LL} & \textbf{CL} & \textbf{CC} & \textbf{CR} & \textbf{RR} & \textbf{FR} & \textbf{BT} \\ \hline

\multirow{9}{*}{\textbf{Mistral}} & \multirow{9}{*}{French} & \textbf{FL} && 1.0000 & 1.0000 & 1.0000 & 0.9995 & 0.0092 & 0.0000 & 0.4689 \\
 & & \textbf{LL} & 0.0000 & & 0.8222 & 0.7063 & 0.1071 & 0.0000 & 0.0000 & 0.0000 \\
 & & \textbf{CL} & 0.0000 & 0.1780 & & 0.3692 & 0.0157 & 0.0000 & 0.0000 & 0.0000 \\
 & & \textbf{CC} & 0.0000 & 0.2940 & 0.6312 & & 0.0369 & 0.0000 & 0.0000 & 0.0000 \\
 & & \textbf{CR} & 0.0005 & 0.8931 & 0.9844 & 0.9632 & & 0.0000 & 0.0000 & 0.0018 \\
 & & \textbf{RR} & 0.9908 & 1.0000 & 1.0000 & 1.0000 & 1.0000 & & 0.0017 & 0.9836 \\
 & & \textbf{FR} & 1.0000 & 1.0000 & 1.0000 & 1.0000 & 1.0000 & 0.9983 & & 1.0000 \\
 & & \textbf{BT} & 0.5318 & 1.0000 & 1.0000 & 1.0000 & 0.9982 & 0.0164 & 0.0000 & \\ \hline

\multirow{9}{*}{\textbf{Mistral}} & \multirow{9}{*}{English} & \textbf{FL} && 1.0000 & 1.0000 & 1.0000 & 1.0000 & 0.0111 & 0.0000 & 0.2251 \\
 & & \textbf{LL} & 0.0000 & & 0.9663 & 0.8427 & 0.3880 & 0.0000 & 0.0000 & 0.0000 \\
 & & \textbf{CL} & 0.0000 & 0.0338 & & 0.1916 & 0.0152 & 0.0000 & 0.0000 & 0.0000 \\
 & & \textbf{CC} & 0.0000 & 0.1575 & 0.8086 & & 0.1010 & 0.0000 & 0.0000 & 0.0000 \\
 & & \textbf{CR} & 0.0000 & 0.6124 & 0.9848 & 0.8992 & & 0.0000 & 0.0000 & 0.0000 \\
 & & \textbf{RR} & 0.9889 & 1.0000 & 1.0000 & 1.0000 & 1.0000 & & 0.0034 & 0.9217 \\
 & & \textbf{FR} & 1.0000 & 1.0000 & 1.0000 & 1.0000 & 1.0000 & 0.9966 & & 0.9998 \\
 & & \textbf{BT} & 0.7754 & 1.0000 & 1.0000 & 1.0000 & 1.0000 & 0.0786 & 0.0002 & \\ \hline

\multirow{9}{*}{\textbf{Mistral}} & \multirow{9}{*}{Spanish} & \textbf{FL} && 0.9999 & 1.0000 & 1.0000 & 0.9996 & 0.0402 & 0.0000 & 0.5877 \\
 & & \textbf{LL} & 0.0001 & & 0.9329 & 0.7238 & 0.3529 & 0.0000 & 0.0000 & 0.0006 \\
 & & \textbf{CL} & 0.0000 & 0.0672 & & 0.1979 & 0.0283 & 0.0000 & 0.0000 & 0.0000 \\
 & & \textbf{CC} & 0.0000 & 0.2766 & 0.8023 & & 0.1760 & 0.0000 & 0.0000 & 0.0001 \\
 & & \textbf{CR} & 0.0004 & 0.6475 & 0.9717 & 0.8243 & & 0.0000 & 0.0000 & 0.0048 \\
 & & \textbf{RR} & 0.9599 & 1.0000 & 1.0000 & 1.0000 & 1.0000 & & 0.0015 & 0.9683 \\
 & & \textbf{FR} & 1.0000 & 1.0000 & 1.0000 & 1.0000 & 1.0000 & 0.9985 & & 1.0000 \\
 & & \textbf{BT} & 0.4129 & 0.9994 & 1.0000 & 0.9999 & 0.9952 & 0.0318 & 0.0000 & \\ \hline

\multirow{9}{*}{\textbf{Mistral}} & \multirow{9}{*}{Russian} & \textbf{FL} && 0.9329 & 0.9976 & 0.9941 & 0.9981 & 0.2543 & 0.0157 & 0.0625 \\
 & & \textbf{LL} & 0.0672 & & 0.9274 & 0.8660 & 0.9453 & 0.0159 & 0.0001 & 0.0016 \\
 & & \textbf{CL} & 0.0024 & 0.0727 & & 0.3936 & 0.5800 & 0.0003 & 0.0000 & 0.0000 \\
 & & \textbf{CC} & 0.0060 & 0.1342 & 0.6068 & & 0.6981 & 0.0010 & 0.0000 & 0.0001 \\
 & & \textbf{CR} & 0.0019 & 0.0548 & 0.4204 & 0.3022 & & 0.0003 & 0.0000 & 0.0000 \\
 & & \textbf{RR} & 0.7461 & 0.9841 & 0.9997 & 0.9990 & 0.9997 & & 0.0706 & 0.1612 \\
 & & \textbf{FR} & 0.9844 & 0.9999 & 1.0000 & 1.0000 & 1.0000 & 0.9296 & & 0.6734 \\
 & & \textbf{BT} & 0.9377 & 0.9984 & 1.0000 & 0.9999 & 1.0000 & 0.8393 & 0.3272 & \\ \hline

\multirow{9}{*}{\textbf{Mistral}} & \multirow{9}{*}{Chinese} & \textbf{FL} && 0.9663 & 0.9816 & 0.9823 & 0.9769 & 0.9411 & 0.9215 & 0.8815 \\
 & & \textbf{LL} & 0.0338 & & 0.6971 & 0.6363 & 0.6616 & 0.4373 & 0.3855 & 0.3559 \\
 & & \textbf{CL} & 0.0184 & 0.3033 & & 0.4719 & 0.4927 & 0.3160 & 0.2562 & 0.2514 \\
 & & \textbf{CC} & 0.0177 & 0.3641 & 0.5285 & & 0.4868 & 0.3150 & 0.2512 & 0.2562 \\
 & & \textbf{CR} & 0.0231 & 0.3387 & 0.5077 & 0.5136 & & 0.3316 & 0.2547 & 0.2664 \\
 & & \textbf{RR} & 0.0590 & 0.5631 & 0.6844 & 0.6854 & 0.6688 & & 0.4115 & 0.4165 \\
 & & \textbf{FR} & 0.0786 & 0.6149 & 0.7442 & 0.7492 & 0.7457 & 0.5889 & & 0.4833 \\
 & & \textbf{BT} & 0.1188 & 0.6447 & 0.7490 & 0.7443 & 0.7341 & 0.5842 & 0.5173 & \\ \hline

\multirow{9}{*}{\textbf{Mistral}} & \multirow{9}{*}{Arabic} & \textbf{FL} && 0.5476 & 0.9727 & 0.7942 & 0.7076 & 0.5720 & 0.4299 & 0.4309 \\
 & & \textbf{LL} & 0.4528 & & 0.9685 & 0.7800 & 0.6483 & 0.5279 & 0.3892 & 0.4311 \\
 & & \textbf{CL} & 0.0274 & 0.0316 & & 0.1394 & 0.0715 & 0.0403 & 0.0209 & 0.0382 \\
 & & \textbf{CC} & 0.2061 & 0.2203 & 0.8608 & & 0.3720 & 0.2468 & 0.1598 & 0.2193 \\
 & & \textbf{CR} & 0.2928 & 0.3521 & 0.9286 & 0.6284 & & 0.3655 & 0.2540 & 0.2662 \\
 & & \textbf{RR} & 0.4284 & 0.4725 & 0.9598 & 0.7535 & 0.6349 & & 0.3626 & 0.3841 \\
 & & \textbf{FR} & 0.5705 & 0.6112 & 0.9792 & 0.8405 & 0.7464 & 0.6379 & & 0.5096 \\
 & & \textbf{BT} & 0.5698 & 0.5695 & 0.9619 & 0.7811 & 0.7343 & 0.6165 & 0.4910 & \\ \hline

\end{tabular}
\end{table*}

\begin{table*}
\centering
\caption{Qwen-72B: Mann-Withney test p-values.}
\label{tab:qwen72b}
\begin{tabular}{l|l|ccccccccc}
\hline
\textbf{Model} & \textbf{Language} &  & \textbf{FL} & \textbf{LL} & \textbf{CL} & \textbf{CC} & \textbf{CR} & \textbf{RR} & \textbf{FR} & \textbf{BT} \\ \hline

\multirow{9}{*}{\textbf{Qwen72b}} & \multirow{9}{*}{French} & \textbf{FL} & &1.0000	&1.0000	&0.9999	&0.9606	&0.0012	&0.0000	&0.7133
\\
 & & \textbf{LL} & 0.0000&		&0.9502	&0.1348	&0.0010	&0.0000	&0.0000	&0.0001\\
 & & \textbf{CL} &0.0000	&0.0498	&	&0.0031	&0.0000	&0.0000	&0.0000	&0.0000 \\
 & & \textbf{CC} & 0.0001&	0.8654	&0.9969	&	&0.0255	&0.0000	&0.0000	&0.0029 \\
 & & \textbf{CR} & 0.0395&	0.9990	&1.0000	&0.9746	&	&0.0000	&0.0000	&0.1545\\
 & & \textbf{RR} &0.9988	&1.0000	&1.0000	&1.0000	&1.0000	&	&0.0002	&0.9995\\
 & & \textbf{FR} &1.0000	&1.0000	&1.0000	&1.0000	&1.0000	&0.9998	&	&1.0000\\
 & & \textbf{BT} &0.2872	&0.9999	&1.0000	&0.9971	&0.8459	&0.0005	&0.0000
&  \\ \hline

\multirow{9}{*}{\textbf{Qwen72b}} & \multirow{9}{*}{English} & \textbf{FL} & &1.0000	&1.0000	&1.0000	&0.9872	&0.0350	&0.0000	&0.7508
\\
 & & \textbf{LL} & 0.0000&		&0.9378	&0.2113	&0.0023	&0.0000	&0.0000	&0.0001\\
 & & \textbf{CL} &0.0000	&0.0623	&	&0.0088	&0.0000	&0.0000	&0.0000	&0.0000 \\
 & & \textbf{CC} & 0.0000&	0.7890	&0.9912	&	&0.0246	&0.0000	&0.0000	&0.0012 \\
 & & \textbf{CR} & 0.0128&	0.9977	&1.0000	&0.9754	&	&0.0000	&0.0000	&0.0924\\
 & & \textbf{RR} &0.9651	&1.0000	&1.0000	&1.0000	&1.0000	&	&0.0000	&0.9881\\
 & & \textbf{FR} &1.0000	&1.0000	&1.0000	&1.0000	&1.0000	&1.0000	&	&1.0000\\
 & & \textbf{BT} &0.2498	&0.9999	&1.0000	&0.9988	&0.9079	&0.0120	&0.0000
&  \\ \hline

\multirow{9}{*}{\textbf{Qwen72b}} & \multirow{9}{*}{Spanish} & \textbf{FL} & &1.0000	&1.0000	&0.9999	&0.9458	&0.0098	&0.0000	&0.6032
\\
 & & \textbf{LL} & 0.0000&		&0.8690	&0.1832	&0.0008	&0.0000	&0.0000	&0.0001\\
 & & \textbf{CL} &0.0000	&0.1312	&	&0.0201	&0.0000	&0.0000	&0.0000	&0.0000 \\
 & & \textbf{CC} & 0.0001&	0.8171	&0.9800	&	&0.0054	&0.0000	&0.0000	&0.0005 \\
 & & \textbf{CR} & 0.0543&	0.9992	&1.0000	&0.9946	&	&0.0000	&0.0000	&0.0905\\
 & & \textbf{RR} &0.9902	&1.0000	&1.0000	&1.0000	&1.0000	&	&0.0000	&0.9899\\
 & & \textbf{FR} &1.0000	&1.0000	&1.0000	&1.0000	&1.0000	&1.0000	&	&1.0000\\
 & & \textbf{BT} &0.3975	&0.9999	&1.0000	&0.9995	&0.9098	&0.0101	&0.0000
&  \\ \hline

\multirow{9}{*}{\textbf{Qwen72b}} & \multirow{9}{*}{Russian} & \textbf{FL} && 1.0000 & 1.0000 & 1.0000 & 0.9997 & 0.3191 & 0.0001 & 0.7784 \\
 & & \textbf{LL} & 0.0000 & & 0.9970 & 0.7313 & 0.2498 & 0.0000 & 0.0000 & 0.0033 \\
 & & \textbf{CL} & 0.0000 & 0.0030 & & 0.0136 & 0.0003 & 0.0000 & 0.0000 & 0.0000 \\
 & & \textbf{CC} & 0.0000 & 0.2690 & 0.9865 & & 0.1139 & 0.0000 & 0.0000 & 0.0009 \\
 & & \textbf{CR} & 0.0003 & 0.7505 & 0.9997 & 0.8863 & & 0.0000 & 0.0000 & 0.0113 \\
 & & \textbf{RR} & 0.6813 & 1.0000 & 1.0000 & 1.0000 & 1.0000 & & 0.0004 & 0.8899 \\
 & & \textbf{FR} & 0.9999 & 1.0000 & 1.0000 & 1.0000 & 1.0000 & 0.9996 & & 1.0000 \\
 & & \textbf{BT} & 0.2221 & 0.9967 & 1.0000 & 0.9991 & 0.9887 & 0.1104 & 0.0000 & \\ \hline

\multirow{9}{*}{\textbf{Qwen72b}} & \multirow{9}{*}{Chinese} & \textbf{FL} && 0.9998 & 1.0000 & 0.9823 & 0.9156 & 0.0060 & 0.0000 & 0.4621 \\
 & & \textbf{LL} & 0.0002 & & 0.8424 & 0.0405 & 0.0113 & 0.0000 & 0.0000 & 0.0006 \\
 & & \textbf{CL} & 0.0000 & 0.1578 & & 0.0060 & 0.0008 & 0.0000 & 0.0000 & 0.0000 \\
 & & \textbf{CC} & 0.0177 & 0.9596 & 0.9940 & & 0.2404 & 0.0000 & 0.0000 & 0.0246 \\
 & & \textbf{CR} & 0.0846 & 0.9887 & 0.9992 & 0.7599 & & 0.0000 & 0.0000 & 0.0979 \\
 & & \textbf{RR} & 0.9940 & 1.0000 & 1.0000 & 1.0000 & 1.0000 & & 0.0190 & 0.9816 \\
 & & \textbf{FR} & 1.0000 & 1.0000 & 1.0000 & 1.0000 & 1.0000 & 0.9810 & & 0.9999 \\
 & & \textbf{BT} & 0.5386 & 0.9994 & 1.0000 & 0.9755 & 0.9024 & 0.0185 & 0.0001 & \\ \hline

\multirow{9}{*}{\textbf{Qwen72b}} & \multirow{9}{*}{Arabic} & \textbf{FL} & & 0.9990 & 1.0000 & 0.7742 & 0.4334 & 0.0009 & 0.0000 & 0.3281 \\
 & & \textbf{LL} & 0.0010 & & 0.9193 & 0.0092 & 0.0004 & 0.0000 & 0.0000 & 0.0003 \\
 & & \textbf{CL} & 0.0000 & 0.0809 & & 0.0001 & 0.0000 & 0.0000 & 0.0000 & 0.0000 \\
 & & \textbf{CC} & 0.2261 & 0.9908 & 0.9999 & & 0.1601 & 0.0000 & 0.0000 & 0.1121 \\
 & & \textbf{CR} & 0.5671 & 0.9996 & 1.0000 & 0.8402 & & 0.0010 & 0.0000 & 0.3980 \\
 & & \textbf{RR} & 0.9991 & 1.0000 & 1.0000 & 1.0000 & 0.9990 & & 0.0064 & 0.9918 \\
 & & \textbf{FR} & 1.0000 & 1.0000 & 1.0000 & 1.0000 & 1.0000 & 0.9936 & & 1.0000 \\
 & & \textbf{BT} & 0.6725 & 0.9997 & 1.0000 & 0.8881 & 0.6027 & 0.0082 & 0.0000 & \\ \hline

\end{tabular}
\end{table*}

\begin{table*}
\centering
\caption{Qwen-7B: Mann-Withney test p-values.}
\label{tab:qwen7b}
\begin{tabular}{l|l|ccccccccc}
\hline
\textbf{Model} & \textbf{Language} &  & \textbf{FL} & \textbf{LL} & \textbf{CL} & \textbf{CC} & \textbf{CR} & \textbf{RR} & \textbf{FR} & \textbf{BT} \\ \hline
\multirow{9}{*}{\textbf{Qwen7b}} & \multirow{9}{*}{French} & \textbf{FL} & &0.9997 & 0.9996 & 0.9825 & 0.1835 & 0.0011 & 0.0000 & 0.6691 \\
 & & \textbf{LL} & 0.0003 & & 0.4588 & 0.0397 & 0.0000 & 0.0000 & 0.0000 & 0.0051 \\
 & & \textbf{CL} & 0.0004 & 0.5415 & & 0.0716 & 0.0000 & 0.0000 & 0.0000 & 0.0075 \\
 & & \textbf{CC} & 0.0175 & 0.9604 & 0.9286 & & 0.0007 & 0.0000 & 0.0000 & 0.0964 \\
 & & \textbf{CR} & 0.8168 & 1.0000 & 1.0000 & 0.9993 & & 0.0158 & 0.0003 & 0.9051 \\
 & & \textbf{RR} & 0.9989 & 1.0000 & 1.0000 & 1.0000 & 0.9843 & & 0.0659 & 0.9991 \\
 & & \textbf{FR} & 1.0000 & 1.0000 & 1.0000 & 1.0000 & 0.9997 & 0.9342 & & 1.0000 \\
 & & \textbf{BT} & 0.3315 & 0.9949 & 0.9925 & 0.9039 & 0.0952 & 0.0009 & 0.0000 & \\ \hline

\multirow{9}{*}{\textbf{Qwen7b}} & \multirow{9}{*}{English} & \textbf{FL} & &0.9954 & 0.9988 & 0.9736 & 0.5031 & 0.0065 & 0.0000 & 0.7572 \\
 & & \textbf{LL} & 0.0046 & & 0.6054 & 0.1748 & 0.0034 & 0.0000 & 0.0000 & 0.0470 \\
 & & \textbf{CL} & 0.0012 & 0.3949 & & 0.0911 & 0.0007 & 0.0000 & 0.0000 & 0.0233 \\
 & & \textbf{CC} & 0.0264 & 0.8255 & 0.9090 & & 0.0213 & 0.0000 & 0.0000 & 0.1673 \\
 & & \textbf{CR} & 0.4974 & 0.9966 & 0.9993 & 0.9788 & & 0.0038 & 0.0000 & 0.7376 \\
 & & \textbf{RR} & 0.9936 & 1.0000 & 1.0000 & 1.0000 & 0.9962 & & 0.0081 & 0.9985 \\
 & & \textbf{FR} & 1.0000 & 1.0000 & 1.0000 & 1.0000 & 1.0000 & 0.9919 & & 1.0000 \\
 & & \textbf{BT} & 0.2433 & 0.9532 & 0.9768 & 0.8331 & 0.2630 & 0.0016 & 0.0000 & \\ \hline

\multirow{9}{*}{\textbf{Qwen7b}} & \multirow{9}{*}{Spanish} & \textbf{FL} & &0.9796 & 0.9621 & 0.9854 & 0.3345 & 0.0153 & 0.0001 & 0.3945 \\
 & & \textbf{LL} & 0.0204 & & 0.3828 & 0.5258 & 0.0047 & 0.0000 & 0.0000 & 0.0298 \\
 & & \textbf{CL} & 0.0380 & 0.6175 & & 0.6566 & 0.0093 & 0.0000 & 0.0000 & 0.0413 \\
 & & \textbf{CC} & 0.0146 & 0.4746 & 0.3438 & & 0.0031 & 0.0000 & 0.0000 & 0.0235 \\
 & & \textbf{CR} & 0.6659 & 0.9953 & 0.9907 & 0.9969 & & 0.0491 & 0.0008 & 0.6110 \\
 & & \textbf{RR} & 0.9847 & 1.0000 & 1.0000 & 1.0000 & 0.9510 & & 0.0563 & 0.9547 \\
 & & \textbf{FR} & 0.9999 & 1.0000 & 1.0000 & 1.0000 & 0.9992 & 0.9439 & & 0.9984 \\
 & & \textbf{BT} & 0.6061 & 0.9703 & 0.9588 & 0.9765 & 0.3896 & 0.0454 & 0.0016 & \\ \hline

\multirow{9}{*}{\textbf{Qwen7b}} & \multirow{9}{*}{Russian} & \textbf{FL} & &0.9999 & 0.9998 & 0.9987 & 0.9614 & 0.9479 & 0.0020 & 0.9426 \\
 & & \textbf{LL} & 0.0001 & & 0.3918 & 0.1805 & 0.0189 & 0.0053 & 0.0000 & 0.0483 \\
 & & \textbf{CL} & 0.0002 & 0.6085 & & 0.2855 & 0.0401 & 0.0181 & 0.0000 & 0.0730 \\
 & & \textbf{CC} & 0.0013 & 0.8198 & 0.7148 & & 0.1292 & 0.0759 & 0.0000 & 0.1673 \\
 & & \textbf{CR} & 0.0387 & 0.9811 & 0.9600 & 0.8710 & & 0.3805 & 0.0000 & 0.5160 \\
 & & \textbf{RR} & 0.0522 & 0.9947 & 0.9819 & 0.9242 & 0.6200 & & 0.0000 & 0.6268 \\
 & & \textbf{FR} & 0.9980 & 1.0000 & 1.0000 & 1.0000 & 1.0000 & 1.0000 & & 1.0000 \\
 & & \textbf{BT} & 0.0576 & 0.9518 & 0.9272 & 0.8331 & 0.4846 & 0.3739 & 0.0000 & \\ \hline

\multirow{9}{*}{\textbf{Qwen7b}} & \multirow{9}{*}{Chinese} & \textbf{FL} & &0.9418 & 0.9783 & 0.9811 & 0.5181 & 0.1145 & 0.0677 & 0.5628 \\
 & & \textbf{LL} & 0.0583 & & 0.6840 & 0.7046 & 0.0633 & 0.0019 & 0.0010 & 0.1146 \\
 & & \textbf{CL} & 0.0218 & 0.3164 & & 0.5337 & 0.0256 & 0.0004 & 0.0002 & 0.0663 \\
 & & \textbf{CC} & 0.0190 & 0.2958 & 0.4667 & & 0.0243 & 0.0004 & 0.0003 & 0.0620 \\
 & & \textbf{CR} & 0.4823 & 0.9369 & 0.9744 & 0.9758 & & 0.0994 & 0.0616 & 0.5370 \\
 & & \textbf{RR} & 0.8858 & 0.9981 & 0.9996 & 0.9996 & 0.9008 & & 0.3757 & 0.9073 \\
 & & \textbf{FR} & 0.9324 & 0.9990 & 0.9998 & 0.9997 & 0.9385 & 0.6247 & & 0.9218 \\
 & & \textbf{BT} & 0.4379 & 0.8857 & 0.9339 & 0.9381 & 0.4637 & 0.0930 & 0.0784 & \\ \hline

\multirow{9}{*}{\textbf{Qwen7b}} & \multirow{9}{*}{Arabic} & \textbf{FL} & &0.5893 & 0.9606 & 0.8169 & 0.7627 & 0.4780 & 0.3857 & 0.3543 \\
 & & \textbf{LL} & 0.4111 & & 0.9580 & 0.7370 & 0.6596 & 0.3794 & 0.2839 & 0.2796 \\
 & & \textbf{CL} & 0.0394 & 0.0421 & & 0.1684 & 0.1345 & 0.0260 & 0.0189 & 0.0241 \\
 & & \textbf{CC} & 0.1834 & 0.2633 & 0.8318 & & 0.4354 & 0.1693 & 0.1140 & 0.1290 \\
 & & \textbf{CR} & 0.2377 & 0.3408 & 0.8658 & 0.5651 & & 0.2177 & 0.1608 & 0.1835 \\
 & & \textbf{RR} & 0.5224 & 0.6210 & 0.9740 & 0.8310 & 0.7827 & & 0.3768 & 0.3911 \\
 & & \textbf{FR} & 0.6147 & 0.7164 & 0.9812 & 0.8862 & 0.8395 & 0.6237 & & 0.4811 \\
 & & \textbf{BT} & 0.6463 & 0.7209 & 0.9760 & 0.8713 & 0.8169 & 0.6096 & 0.5196 & \\ \hline

\end{tabular}
\end{table*}

\section{Computing Infrastructure}

The simulations were conducted using NVIDIA A100 GPUs. Each simulation per language took approximately 24 hours to complete. Smaller models, such as Llama 3-8B, Qwen-7B, and Mistral-7B, were run on a single GPU due to their lower computational demands. In contrast, larger models required four GPUs to handle their increased complexity and resource requirements. For GPT-4o-mini, we utilized the Batch API, sending each prompt as a single request to streamline processing and ensure efficient handling of the workload.



\end{document}